\documentclass[11pt]{article}

\usepackage[margin=1in]{geometry}
\usepackage[utf8]{inputenc}
\usepackage[T1]{fontenc}

\usepackage{amsmath}
\usepackage{amssymb}
\usepackage{amsthm}
\usepackage{amsfonts}

\usepackage{graphicx}
\graphicspath{{figures/}}
\usepackage{float}
\usepackage{placeins}
\usepackage{subcaption}
\usepackage{wrapfig}

\usepackage{booktabs}
\usepackage{multirow}
\usepackage{multicol}
\usepackage{array}
\usepackage{diagbox}
\usepackage{adjustbox}

\usepackage{algorithm}
\usepackage[noend]{algpseudocode}

\usepackage[dvipsnames]{xcolor}
\usepackage[hidelinks]{hyperref}
\usepackage{url}

\usepackage[numbers,sort&compress]{natbib}

\usepackage{authblk}

\usepackage{xspace}
\usepackage{listings}
\usepackage{textcomp}
\usepackage{tcolorbox}
\usepackage{pifont}
\usepackage{breakcites}
\usepackage{enumitem}
\usepackage{bbding}
\usepackage{framed}
\usepackage{annotate-equations}
\definecolor{shadecolor}{rgb}{0.92,0.92,0.92}

\providecommand{\keywords}[1]{\par\noindent\textbf{Keywords:} #1}

\newcommand{\Approach}[1]{{PIMbot}}

\title{\textbf{\Approach{}: A Self-Adaptive Attack Framework for Adversarial Manipulation of Multi-Robot Reinforcement Learning}}

\author[1]{Zexin Li}
\author[1]{Ziliang Zhang}
\author[1]{Hyoseung Kim}
\author[1]{Cong Liu}
\affil[1]{University of California, Riverside\\\texttt{\{zli536,zzhan357\}@ucr.edu, \{hyoseung,congl\}@ucr.edu}}
\date{}

\begin{document}
\maketitle

\begin{abstract}
Recent research has demonstrated the potential of reinforcement learning in effective multi-robot collaboration, particularly in social dilemmas where robots face a trade-off between self-interest and collective benefits. However, environmental factors such as miscommunication and adversarial robots can impact cooperation, making it crucial to explore how multi-robot communication can be manipulated to achieve different outcomes. This paper presents \Approach{}, a framework that manipulates outcomes via two complementary levers: (i) incentive manipulation of the reward channel and (ii) policy manipulation of an agent's own actions. An adaptive multi-objective controller balances these levers in an online manner. Our work introduces a novel approach to manipulation in recent multi-agent RL social dilemmas that utilize a unique reward function for incentivization. By utilizing our proposed PIMbot mechanisms, a robot is able to manipulate the social dilemma environment effectively. Comprehensive experimental results demonstrate the effectiveness of our proposed methods in the Gazebo-simulated multi-robot environment. Moreover, a real embedded device case study on NVIDIA Jetson Orin Nano quantifies systems cost and validates \Approach{}'s effectiveness on realistic autonomous embedded systems scenarios beyond simulation. Together, these results position \Approach{} as a rigorous stress-test tool exposing critical vulnerabilities in multi-robot cooperative tasks.
\end{abstract}

\keywords{Multi-Robot Systems, Reinforcement Learning, Adversarial Attacks, Self-Adaptive Manipulation, Social Dilemmas, Multi-Agent Coordination}

\section{Introduction}

Intelligent robots have been widely used in various applications, ranging from pick-and-place operations in manufacturing to assisting with disorder detection in healthcare settings~\cite{ahangar2019design, su15107790}. Despite these advances, challenges persist when robots must operate in complex, dynamic, and unpredictable environments. Multi-robot collaboration offers a promising pathway to overcome these challenges by enabling robots to coordinate, share information, and exploit complementary capabilities. Such collaboration can unlock higher efficiency, robustness, and adaptability compared to isolated robot operation. 
Collaboration scenarios are diverse and domain-specific, involving principled methods for multi-robot decision-making. In manufacturing, multiple robotic arms jointly assemble intricate products with high precision; in warehouse logistics, fleets of autonomous robots can coordinate to optimize task allocation and traffic management; in disaster response, heterogeneous robots such as aerial drones and ground vehicles can cooperate to search, rescue, and deliver supplies; and in space exploration, excavation, construction, and inspection robots must work under harsh conditions with sparse feedback and delayed communication.

Reinforcement learning (RL), a class of trial-and-error learning algorithms, has emerged as a powerful approach for adaptive robot~\cite{TereshchukSBPDB19,AgrawalABM22iros,GaoWZYWXWLXG22iros,ZhangQQXWZWC0ZL22iros}, as well as control, and security fields~\cite{guo2023backdoor,BaiZLZ21iros,siedler2022dynamic,zhang2020decentralized}. Agents adopt RL to learn policies that maximize long-term rewards by interacting with their environments, making it effective in high-dimensional and uncertain domains. Extending RL to multi-agent settings, known as Multi-Agent Reinforcement Learning (MARL), provides a strong theoretical and algorithmic foundation for coordinating multiple robots in cooperative, competitive operations while adapting to the behaviors of others. Consequently, MARL is particularly fitting for multi-robot systems that address challenges beyond the single-robot solutions. In the context of multi-robot systems, RL has demonstrated clear benefits in practical scenarios such as reforestation using coordinated swarms~\cite{siedler2022dynamic}, cooperative transport of large objects~\cite{zhang2020decentralized}, and large-scale exploration in unknown environments~\cite{GaoWZYWXWLXG22iros}. Despite these advances, multi-agent settings often encounter \emph{social dilemmas}, where individual incentives may conflict with collective goals, leading to suboptimal group performance~\cite{stimpson2003learning,yu2020distributed}. Various approaches have been proposed to address these challenges, including game-theoretic reward shaping, communication protocols, and incentive design~\cite{10.5555/3237383.3237408,10.5555/3495724.3496999,han2022solution,he2022robust,li2025attacking,andam2025constrained,yang2021adaptive,zhou2024reciprocal,yu2024robust}. Notably, frameworks such as Learned Incentive Optimization (LIO)~\cite{10.5555/3495724.3496999} highlight the potential of adaptive incentive mechanisms in aligning agent behaviors with global objectives, thereby advancing cooperation in complex multi-robot environments.

Although inter-robot communication can partially alleviate social dilemmas by enabling coordination and information sharing, it is far from a panacea. In practice, the presence of self-interested or even adversarial agents can still derail cooperation, undermining both system efficiency and robustness. This raises a fundamental question: \emph{How can an agent manipulate inter-robot communication and action choices to steer outcomes?}  To address this question, we introduce \Approach{}, a novel framework that explicitly models manipulation in multi-robot reinforcement learning systems. Our preliminary work~\cite{nikkhoo2023pimbot} is built around two complementary levers: (\textit{i}) \emph{incentive manipulation}, which targets the reward channel to reshape agent preferences, and (\textit{ii}) \emph{policy manipulation}, which directly alters environment actions to bias interaction outcomes. By combining these levers, our framework provides a unified view of how external or internal agents can shape coordination dynamics. Based on these strategy-based methods, importantly, we propose an \emph{adaptive multi-objective optimization} (ADMO) that tunes these levers online to achieve a desired adversarial outcome mode. This adaptive mechanism allows \Approach{} to flexibly shift between stealthy self-maximization and active team disruption strategies, depending on the overarching system goals and observed agent behaviors. We rigorously validate \Approach{} across two canonical benchmarks, the Escape Room (ER) and the Iterated Prisoner’s Dilemma (IPD). Beyond abstract benchmarks, we demonstrate the practicality of \Approach{} in the Gazebo robotic simulator, where multiple robots must collaborate under feedback. Moreover, to highlight real-world feasibility, we outline a case study on physical deployment using an NVIDIA Jetson Orin Nano, providing realistic insights into on-device constraints and opportunities for applying manipulation-aware MARL in real robotic systems. 

Our major contributions are:
\begin{itemize}
\item We provide a formalization of two key manipulation dimensions in multi-robot social dilemmas: (\textit{i}) \emph{policy manipulation}, which directly alters action choices in the environment, and (\textit{ii}) \emph{incentive reward manipulation}, which reshapes payoff structures to bias agent behaviors. Building on this, we introduce an \emph{adaptive multi-objective balancing} scheme that jointly tunes these levers and yields Pareto-stationary updates, ensuring stable trade-offs between maximizing the adversary's individual reward and actively disrupting the team's collective success.  

\item We conduct a comprehensive empirical study on the effectiveness of \Approach{}. Specifically, we demonstrate that it can degrade task success rates when adversarial manipulation is applied. These outcomes are validated in canonical multi-agent benchmarks, i.e., the Escape Room (ER) and the Iterated Prisoner’s Dilemma (IPD), and further extended to realistic robotic collaboration tasks within the Gazebo simulator.  

\item {We validate the effectiveness of the proposed method via a realistic case study on a real embedded platform (\emph{NVIDIA Jetson Orin Nano}), providing on-device scenarios that facilitate real-world deployment analysis. This hardware-level validation bridges the gap between simulation and practice, illustrating the feasibility of manipulation-aware MARL in resource-constrained robotic platforms and highlighting potential deployment pathways for future multi-robot systems. To foster reproducibility and community adoption, we have open-sourced the \Approach{} codebase together with experiment configurations.\footnote{\url{https://github.com/UCR-Intelligent-Robotics-Lab/greedy_agent}} }
\end{itemize}

\section{Background}

In this section, we provide an overview of how multi-agent reinforcement learning (MARL) can facilitate effective collaboration among multiple robots. We begin with the basic formulation of reinforcement learning (RL) using policy gradients, which serves as the foundation for extending learning to multi-agent systems. We then discuss how these principles are adapted to settings where multiple autonomous agents must coordinate or compete, and why these extensions are critical for addressing multi-robot social dilemmas.

\subsection{Multi-agent Reinforcement Learning based on Policy Gradient}

At its core, reinforcement learning models the interaction between an agent and its environment as a sequential decision-making problem. In an episodic, discrete-action setting, at time $t$ the agent observes a state $s_t \!\in\! S$, selects an action $a_t \!\in\! A$ according to its policy $\pi_\theta$, and receives an immediate reward $r_t$. The overarching goal is to learn a policy that maximizes the expected cumulative discounted return:  
\begin{equation}
J(\theta) = \mathbb{E}_\theta \!\left[\sum_{t=0}^{\infty} \gamma^t r_t\right],
\label{eq1}
\end{equation}
where $\gamma \in [0,1)$ is a discount factor that balances the importance of immediate versus long-term rewards. This formulation captures the essence of adaptive behavior, which is an agent must not only optimize for short-term gains but also plan for sustainable performance over time.

To optimize $J(\theta)$, policy gradient methods directly compute the gradient of the objective with respect to the policy parameters:  
\begin{equation}
\nabla_\theta J(\theta) = \mathbb{E}_\theta \!\left[ G(s_t, a_t) \nabla_\theta \log \pi_\theta(a_t \mid s_t) \right],
\label{policyGradient}
\end{equation}
where $G(s_t, a_t) = \sum_{i=t}^{\infty} \gamma^{i-t} r_i$ denotes the return from time step $t$ onward. Intuitively, this expression implies that policy updates are guided by the correlation between the chosen actions and their long-term consequences: actions that lead to higher returns are reinforced, while those leading to poor outcomes are suppressed.

\subsection{Incentivization in Policy Gradient}

Cooperative robotics requires effective incentivization methods to align the behavior of multiple agents toward achieving high collective returns. In multi-agent reinforcement learning (MARL), each agent typically learns from a mixture of \emph{extrinsic} and \emph{intrinsic} rewards. Extrinsic rewards are provided directly by the environment (e.g., task completion or resource acquisition), whereas intrinsic rewards can be designed or learned to reflect additional objectives, such as cooperation, exploration, or communication~\cite{Zheng2018OnLI,li2025attacking,andam2025constrained,yang2021adaptive}. The central challenge is that in cooperative tasks, reward signals are often sparse, delayed, or unevenly distributed among agents. As a result, naïvely optimizing self-interested policies can lead to suboptimal equilibria, where agents pursue individual gains at the expense of team performance.    To overcome this misalignment, incentive mechanisms have been proposed to reshape agents’ reward landscapes. By allowing agents to reward or penalize one another, the system can embed cooperative pressures that encourage convergence toward socially optimal outcomes. This is particularly important in multi-robot collaboration, where communication delays, conflicting goals, or heterogeneous capabilities can otherwise destabilize coordination.   One influential framework in this domain is Learned Incentive Optimization (LIO)~\cite{10.5555/3495724.3496999}. LIO extends policy gradient methods by enabling each agent to learn an \emph{incentive function} that accounts for both the behavior of other agents and its own long-term objectives. In the general case of $N$ agents, each agent $i$ is endowed with its own observation $o^i$, action $a^i$, and incentive function $r_{\eta^i}$, which maps its observation and the actions of others to incentive signals. Here, the notation $i$ refers to the \emph{reward-giving} role of an agent, while $j$ refers to the \emph{reward-receiving} role; $-i$ and $-j$ denote all other givers and receivers, respectively. At each time step $t$, the total reward received by agent $j$ is defined as:  

\begin{equation} \label{lio_reward_func}
r^j\!\left(s_t, \mathbf{a}_t, \eta^{-j}\right):= \underbrace{r^{j, \mathrm{env}}\!\left(s_t, \mathbf{a}_t\right)}_{\text{environment}}+\underbrace{\sum_{i \neq j} r_{\eta^i}^j\!\left(o_t^i, a_t^{-i}\right)}_{\text{incentives}}.
\end{equation}
This formulation captures the interplay between environmental rewards and peer-generated incentives. Importantly, it highlights how inter-agent incentives can reshape local objectives to better align with global team goals. Meanwhile, LIO implements a bi-level optimization scheme. At the lower level, \emph{receivers} update their policies based on the combined reward signal (environmental + incentives). Using Eq.~\eqref{lio_reward_func} and the standard policy gradient (Eq.~\eqref{policyGradient}), the policy of recipient $j$ is updated with step size $\beta$:  

\begin{equation}\label{newtheta}
\hat{\theta}^j \leftarrow \theta^j+\beta \,\nabla _{\theta^j} J^{ \pi }\!\left(\theta^j, \eta^{-j}\right).
\end{equation} 
At the upper level, each \emph{giver} $i$ learns how to adjust its incentive function so as to maximize its long-term objective while considering the induced responses of receivers. The giver update is given by:   
\begin{equation}\label{updateetha}
\nabla _{\eta^i} J^i\!\left(\hat{\tau}^i, \tau^i, \hat{\boldsymbol{\theta}}, \eta^i\right):=\mathbb{E}_{\hat{\boldsymbol{\pi}}}\!\left[\sum_{t=0}^T \gamma^t \hat{r}_t^{i, \text { env }}\right]-\alpha\, L\!\left(\eta^i, \tau^i\right),
\end{equation}
where $L$ represents the budgetary cost of giving incentives, and $\alpha$ controls the trade-off between cooperative shaping and resource expenditure.  

This structure highlights a critical insight: incentivization transforms MARL from a purely self-interested optimization problem into a coupled, hierarchical learning process. By explicitly modeling how one agent’s incentives alter the learning trajectory of others, frameworks like LIO provide a principled mechanism for fostering cooperation. At the same time, they reveal the potential for manipulation: the same incentive channels that promote cooperation can also be exploited by adversarial agents to bias outcomes in their favor. This duality makes incentivization both a powerful tool and a vulnerability in multi-robot systems, an observation that motivates the development of robust manipulation-aware approaches such as \Approach{} in this work.

\section{Design}
\label{sec:design}

\begin{figure}[!tbp]
\centering
\includegraphics[width=0.6\textwidth]{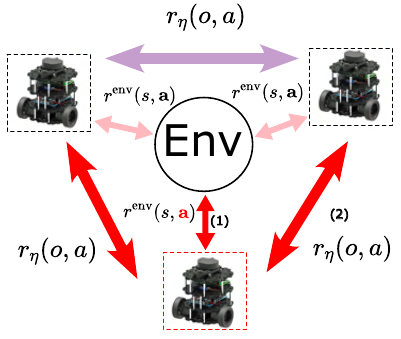}
\caption{A red agent manipulates both the incentive channel (Eq.~\ref{lio_reward_func}) and its policy. By injecting or withholding incentives and by selecting actions strategically, it steers group dynamics.}
\label{bigpic_fig}
\end{figure}

We study incentivized RL for social dilemmas where agents exchange learned incentives and act in a shared environment. We expose two complementary \emph{manipulation levers} (Fig.~\ref{bigpic_fig}): (i) \textbf{incentive manipulation} (tamper with the reward channel) and (ii) \textbf{policy manipulation} (choose actions that steer others).  Furthermore, we propose an adaptive multi-objective controller that balances these levers in an online manner.

\subsection{Incentive Reward Manipulation}
\label{sec:incmanip}

In multi-agent RL with incentives, agents send and receive incentives that shape behavior (Eq.~\ref{lio_reward_func}). This channel is powerful but fragile, and thus a prime target for manipulation. We consider two methods. After that, we propose an adaptive multi-objective manipulation to provide the incorporation of multiple attacking directions.

\noindent\textbf{Partial Communication.}
An agent can act \emph{self-centered}, ignoring the received incentives and optimizing only its extrinsic reward:
\begin{equation} \label{adversaryreward}
r^{SC}\!\left(s_t, \mathbf{a}_t\right) := r^{\mathrm{env}}\!\left(s_t, \mathbf{a}_t\right),
\end{equation}
which changes the update direction of Eq.~\eqref{newtheta} for the adversary (parameters $\theta^{Adv}$). The adversary still \emph{sends} incentives to others (updated by Eq.~\eqref{updateetha}) to influence their policies while remaining immune to their influence. \textit{Why this works.} Because $r^{SC}$ removes the incentive-dependent term from the adversary’s return, the policy gradient $\nabla_{\theta^{Adv}}J^\pi$ in Eq.~\eqref{newtheta} aligns with pure environment returns, while other agents continue to respond to the adversary’s emitted incentives; this asymmetry tilts learning dynamics in favor of the adversary without breaking the communication protocol.

\noindent\textbf{Fake Incentive Reward.}
Without sanity checks on received incentives, an adversary can transmit a constant, oversized incentive $C^{Adv}\!\in\!\mathbb{N}$ that dominates recipients’ immediate returns:
\begin{equation} \label{Cinq}
C^{Adv} \gg r^{\mathrm{env}}\!\left(s_t, \mathbf{a}_t\right)\ \ \text{and}\ \ r_{\eta^i}^j\!\left(o_t^i, a_t^{-i}\right).
\end{equation}
A recipient $j$ then optimizes
\begin{equation} \label{fake_reward}
r^j\!\left(s_t, \mathbf{a}_t, \eta^{-j}\right)= r^{j,\mathrm{env}}\!\left(s_t, \mathbf{a}_t\right)+\!\!\sum_{i \neq j} r_{\eta^i}^j\!\left(o_t^i, a_t^{-i}\right) + C^{Adv},
\end{equation}
which suppresses recipients' exploration by making $\nabla_{\theta^j} J^\pi \!\approx 0$ and thus $\hat{\theta}^j\!\approx\!\theta^j$.

\subsection{Policy Manipulation}
\label{sec:polmanip}

In cooperative multi-robot tasks, coordinated action is critical for success. However, cooperation is not guaranteed: an adversarial agent can manipulate group outcomes purely through its own policy choices, without directly altering rewards or communication channels. This highlights an important complementary attack surface to incentivization, where manipulation arises not from reshaping others’ incentives, but from strategically selecting disruptive actions. Below, we analyze two forms of policy manipulation.  

\noindent\textbf{Bypass Policy (No-Op).}  
One straightforward adversarial strategy is to idle (i.e., perform a no-operation action), thereby refusing to contribute to task execution. Such a bypass policy is particularly disruptive in \emph{non-redundant} tasks that require all agents’ participation to succeed (e.g., ER$(2,1)$, where the aim is to let at least 1 robot push the lever out of 2 robot groups), where even a single idle agent blocks group progress. In contrast, for redundant tasks (e.g., ER$(4,2)$, where the aim is to let at least 2 robots push the lever out of 4 robot groups), the system may still succeed because other agents can compensate (e.g., open the door and exit without the adversary’s help). Formally, if feasibility requires a coalition of size $M^\star$, then idling effectively removes the adversary from all coalitions containing it. The outcome depends on whether the remaining $N{-}M^\star$ benign agents can still form feasible coalitions to complete the task. This illustrates how redundancy provides robustness, but also exposes a clear vulnerability when tasks operate near the minimum coalition threshold.  

\noindent\textbf{Reverse Policy (Reward-Flip).}  
A more aggressive form of manipulation arises when the adversary actively seeks to \emph{invert} the optimization objective. While benign agents maximize  
\begin{equation} \label{PGforgeniun}
\max _{\theta^j} J^{\pi}\!\left(\theta^j, \eta^{-j}\right)=\mathbb{E}_\theta\!\Big[\textstyle\sum_{t=0}^T \gamma^t r^j\!\left(s_t, \mathbf{a}_t, \eta^{-j}\right)\Big],
\end{equation}
an adversary instead minimizes this return by solving $\min_{\theta^{Adv}} J^\pi(\theta^{Adv},\eta^{-Adv})$, equivalently maximizing the negative reward. Conceptually, this transforms the standard policy gradient update (Eq.~\eqref{newtheta}) from gradient ascent to gradient descent, thereby depressing both the environment reward and the incentives that other agents are motivated to provide (see Eq.~\eqref{lio_reward_func}).  

\subsection{{Adaptive Multi-Objective Manipulation}}
\label{sec:MO}

\noindent\textbf{Rationale for loss-based formulation and weighting.}
Sections~\ref{sec:incmanip}–\ref{sec:polmanip} introduced strategy-based manipulations (partial communication, fake incentives, bypass, reward-flip) and their intent: skew a target agent’s margin over others and steer collective welfare up or down. Those strategies are discrete and policy-dependent, which makes their direct optimization non-differentiable and brittle with deep function approximation in MARL. Here we recast the same goals into differentiable surrogates that can be optimized end-to-end with standard policy-gradient machinery. Because multiple strategies may conflict (e.g., an update that strengthens incentive pressure can undo a favorable policy state), we propose a novel ADMO attack, which combines multiple strategies with an adaptive weighted objective with providing a KKT-based auto weight balancing to produce a Pareto-stationary joint descent direction with negligible overhead. 

\noindent\textbf{Incentive-manipulation loss.}
We seek to increase the adversary’s margin over others while respecting an incentive-giving budget. Using only the established reward notation $r$ (and its induced returns), we define 
\begin{equation}
\mathcal{L}_{\mathrm{inc}}
= -\,\mathbb{E}_{\pi_\theta}\!\left[ \sum_{t=0}^{T}\gamma^{t}
\left(
r_t^{a}(s_t,\mathbf{a}_t,\eta^{-a})
- \frac{1}{|\mathcal{N}|-1}\!\!\sum_{j\in\mathcal{N}\setminus\{a\}} \! r_t^{j}(s_t,\mathbf{a}_t,\eta^{-j})
\right)\right]
+ \underbrace{\beta_b L(\eta^{a},\tau)}_{\text{Cost}_\eta}.
\label{eq:inc_loss}
\end{equation}
where $\mathcal{N}=\{1,\dots,N\}$ is the set of agents and  $a\in\mathcal{N}$ is the adversarial agent (so $\theta^{\mathrm{Adv}}{:=}\theta^{a}$, $\eta^{\mathrm{Adv}}{:=}\eta^{a}$).  $L(\eta^{a},\tau)$ is the incentive cost reused from Eq.~\eqref{updateetha}. Intuitively, minimizing $\mathcal{L}_{\mathrm{inc}}$ enlarges the return gap \emph{in favor} of the adversary while penalizing excessive incentive spend via $\beta_b$.
 
\noindent\textbf{Policy-manipulation loss.}
We steer {team welfare}:
\begin{equation}
\label{eq:pol_loss}
\mathcal{L}_{\mathrm{pol}}
= -\, s \cdot \mathcal{W}(\theta,\eta)
\;+\;
\lambda_{d}\, D_{\mathrm{proxy}}\!\big(\theta~\|~\theta_{\mathrm{ref}}\big),
\end{equation}
where $\mathcal{W}(\theta,\eta)=\mathbb{E}_{\pi_\theta}\!\Big[\sum_{t=0}^{T}\gamma^{t}\!\sum_{j\in\mathcal{N}} r_t^{j}(s_t,\mathbf{a}_t,\eta^{-j})\Big]$ denotes team welfare (consistent with Eq.~\eqref{lio_reward_func}). The regularizer
\begin{equation}
\label{eq:proxy_kl}
D_{\mathrm{proxy}}\!\big(\theta~\|~\theta_{\mathrm{ref}}\big)
=\tfrac{1}{2}\sum_{i}(\theta_i-\theta_i^{\mathrm{ref}})^2
\end{equation}
serves as a lightweight surrogate for the KL divergence between $\pi_{\theta}$ and $\pi_{\mathrm{ref}}$, stabilizing optimization by limiting parameter drift.

\noindent\textbf{Weighted-sum objective.}
We minimize a weighted-sum objective function to achieve adaptive combination of multiple strategies.
\begin{equation}
\label{eq:mo_total}
\mathcal{L}_{MO} \;=\; \alpha_1\, \cdot \mathcal{L}_{inc} \;+\; \alpha_2\, \cdot \mathcal{L}_{pol},
\qquad \alpha_1+\alpha_2=1,\ \ \alpha_i\ge c_i.
\end{equation}

\noindent\textbf{Adaptive Pareto weights.}
Let $g_{inc}=\nabla_\omega \mathcal{L}_{inc}$, $g_{pol}=\nabla_\omega \mathcal{L}_{pol}$; $\mathcal{G}=[g_{inc}, g_{pol}]$. We choose weights $\boldsymbol{\alpha}=[\alpha_1,\alpha_2]^\top$ to minimize $\tfrac{1}{2}\|\mathcal{G}\boldsymbol{\alpha}\|_2^2$ subject to Eq.~\eqref{eq:mo_total}. Denote $\boldsymbol{e}=[1,1]^\top$ and $\boldsymbol{c}=[c_1,c_2]^\top$. The KKT closed form yields~\cite{multi-objective-optimization}:
\begin{equation}
\label{eq:alpha_qp}
\begin{gathered}
\begin{bmatrix}
\hat{\boldsymbol{\alpha}}^*\\ 
\lambda
\end{bmatrix}
=\big(\mathcal{M}\mathcal{M}^{\top}\big)^{-1}\mathcal{M}
\begin{bmatrix}
-\mathcal{G}\mathcal{G}^{\top}\boldsymbol{c}\\ 
1-\boldsymbol{e}^{\top}\boldsymbol{c}
\end{bmatrix},
\qquad
\mathcal{M}=
\begin{bmatrix}
\mathcal{G}\mathcal{G}^{\top} & \boldsymbol{e} \\
\boldsymbol{e}^{\top} & 0
\end{bmatrix},
\end{gathered}
\end{equation}
where $\mathcal{G}\in\mathbb{R}^{d\times 2}$ with $d=\dim(\omega)$, and $\mathcal{G}\mathcal{G}^{\top}$ is a rank-$\le 2$ matrix whose entries are gradient inner products. The solution computes the \emph{shortest} common descent direction in the span of $\{g_{inc},g_{pol}\}$. The update direction is
\begin{equation}
\label{eq:pareto_grad}
g^{*}=\hat{\alpha}_1^{*}\, \cdot g_{inc} + \hat{\alpha}_2^{*}\, \cdot g_{pol}. 
\end{equation}

\noindent\textit{Remarks.} (i) The KKT system in Eq.~\eqref{eq:alpha_qp} is $3{\times}3$ and trivial to solve each step; the overhead is negligible relative to policy evaluation. (ii) When $s=-1$ and $c_2$ is large (stalled progress), the controller emphasizes policy-level shaping; when $s=+1$ with large $c_2$ or high incentive variance (large $c_1$), it prefers adversarial pressure. (iii) Bounded incentives and KL regularization make the controller robust to pathological oscillations observed when optimizing the two levers independently.   

\begin{algorithm}[!tbp]
\caption{Adversary with Adaptive Multi-Objective Balancing}
\label{alg:pimbot-mo}
\begin{algorithmic}[1]
\State \textbf{Initialize}: $\omega=[\theta^{a},\eta^{a}]$, $\pi_{\mathrm{ref}}{:=}\pi_{\theta^{a}}$, mode $s\!\in\!\{+1,-1\}$, $(c_1,c_2){=}(0.1,0.1)$
\For{iteration $k=1,2,\dots$}
  \State Roll out trajectories; estimate $J^{a}$, $\bar{J}^{-a}$, $\mathcal{W}$, and $\overline{\mathrm{SR}}$
  \State Update EMAs for variance and success rate
  \State Compute $\mathcal{L}_{\mathrm{inc}}$ (Eq.~\ref{eq:inc_loss}) and $\mathcal{L}_{\mathrm{pol}}$ (Eq.~\ref{eq:pol_loss})
  \State Get gradients $g_{\mathrm{inc}}, g_{\mathrm{pol}}$
  \State Update $(c_1,c_2)$
  \State Solve Eq.~\eqref{eq:alpha_qp} for $\hat{\alpha}^*$
  \State $g^{*}\!\leftarrow\! \hat{\alpha}_1^* g_{\mathrm{inc}} + \hat{\alpha}_2^* g_{\mathrm{pol}}$
  \If{$\|g^{*}\|_2 > 0$}
     \State $\omega \!\leftarrow\! \omega - \eta \cdot \mathrm{AdamW}(g^{*})$
     \State Clip update step to $\|g^{*}\|_2 \leq 1.0$
  \Else
     \State Skip update and retain previous $\omega$
  \EndIf
  \If{$k \bmod K_{\mathrm{ref}} = 0$}
     \State $\pi_{\mathrm{ref}}\!\leftarrow\!\pi_{\theta^{a}}$
     \If{$D_{\mathrm{proxy}}\!\big(\theta~\|~\theta_{\mathrm{ref}}\big) > 0.5$}
        \State Temporarily set $\lambda_d \leftarrow 0.05$
     \EndIf
  \EndIf
  \If{incentive budget $B_{a}$ exceeded}
     \State Zero out $r_{\eta^{a}}$ until next episode
  \EndIf
\EndFor
\end{algorithmic}
\end{algorithm}

\section{Evaluation}
\label{sec:eval}

We evaluate \Approach{} across three representative social-dilemma testbeds: the Escape Room (ER), the Iterated Prisoner Dilemma (IPD), and Stag Hunt (SH), to quantify how \emph{incentive} and \emph{policy} manipulation shape group behavior relative to baselines including Learned Incentive Optimization (LIO)~\cite{10.5555/3495724.3496999} and the Reciprocators framework~\cite{zhou2024reciprocal}. 

\subsection{Experimental Setup}
In this section, we will present the evaluation environment used to test our methods for incentive manipulation in multi-robot reinforcement learning within social dilemmas by manipulating the policy and incentive rewards. We present the results of 10 separate runs for each environment to demonstrate the effectiveness of PIMbot's method.

\subsubsection{Escape Room (ER)}
In our evaluation, we utilized Gazebo to implement the Escape Room environment \cite{10.5555/3495724.3496999}. The Escape Room environment ER($N, M$) is a Markov game for N players with individual extrinsic rewards. To successfully exit the environment, an agent must receive a +10 extrinsic reward by exiting a door. However, the door can only be opened when $M$ other agents cooperate to pull the lever ($M<N$), which incurs an extrinsic penalty of -1 for any movement, thus discouraging all agents from taking cooperative action. We conducted experiments with the cases of ($N=2$, $M=1$), ($N=4$, $M=2$), and ($N=4$, $M=3$) to evaluate the effectiveness of our method. 

\subsubsection{Iterated Prisoner's Dilemma (IPD)}
In addition to our evaluation of the Escape Room environment, we conducted tests on the memory-1 Iterated Prisoner's Dilemma environment using our manipulation method \cite{10.5555/3237383.3237408}. In this environment, each agent observes the joint action taken by themselves and the other agent in the previous round. We used the extrinsic reward table for the environment to evaluate the performance of our method. Table~\ref{table:IPD_reward} specifies the extrinsic rewards for each possible outcome of the environment, including mutual cooperation, mutual defection, and unilateral defection.

\begin{table}[!tbp]
\centering
\caption{IPD reward table}
\label{table:IPD_reward}
\begin{tabular}{c|cc}
\toprule
$\mathrm{Agent} 1 / \mathrm{Agent} 2$ & $\mathrm{Cooperate (C)}$ & $\mathrm{Defect (D)}$ \\
\midrule
$\mathrm{Cooperate (C)}$ & $(-1,-1)$ & $(-3,0)$ \\
$\mathrm{Defect (D)}$ & $(0,-3)$ & $(-2,-2)$ \\
\bottomrule
\end{tabular}
\end{table}

\subsubsection{Stag Hunt}
To further evaluate our method on coordination tasks that demand high trust among agents, we incorporate the Stag Hunt environment across multiple scaling configurations (N=2, 3, and 4 agents). In this classic game-theoretic benchmark, agents simultaneously choose to either hunt a stag (cooperate) or a hare (defect). Successfully hunting a stag yields a high payoff (+5.0) but requires all N agents to cooperate simultaneously on the stag's location; hunting a hare guarantees a smaller payoff (+1.0) regardless of the other agents' actions. This environment is particularly well-suited for evaluating the newly integrated Reciprocators algorithm~\cite{zhou2024reciprocal}, which models reciprocal reward influences to foster cooperation. We define the \emph{Success Rate} in Stag Hunt as the normalized average team reward derived primarily from successful stag captures over an episode (computed as the agent's environmental reward divided by the maximum single-stag payoff of 5.0). During normal cooperative training, this success rate should steadily increase; under ADMO adversarial manipulation, however, the success rate is expected to continuously degrade as the adversary actively breaks the fragile N-agent coordination.

\subsection{Implementation Details}
To enable a fair comparison between our approach and LIO~\cite{10.5555/3495724.3496999}, we used the same hyperparameters $\theta$ and $\eta$ for Eq.\ref{newtheta} and Eq.\ref{updateetha}. We use Python and Tensorflow as a backend following LIO \citep{10.5555/3495724.3496999}. For comparisons against the Reciprocators algorithm~\cite{zhou2024reciprocal}, we implemented our manipulation mechanics within their PyTorch-based framework, aligning network architectures and training hyperparameters to ensure an apples-to-apples evaluation across both the IPD and Stag Hunt domains. For our experiments, we deployed the Escape Room (ER) environment in both ROS and Gazebo for two and four-player versions of the environment. We used the TurtleBot3 simulation platform to demonstrate the behavior of agents in the ER environment. Both the simulation and manipulation were run on an Intel Core i7-10700K CPU.

\begin{figure*}[!tbp]
\centering
\includegraphics[width=0.8\textwidth]{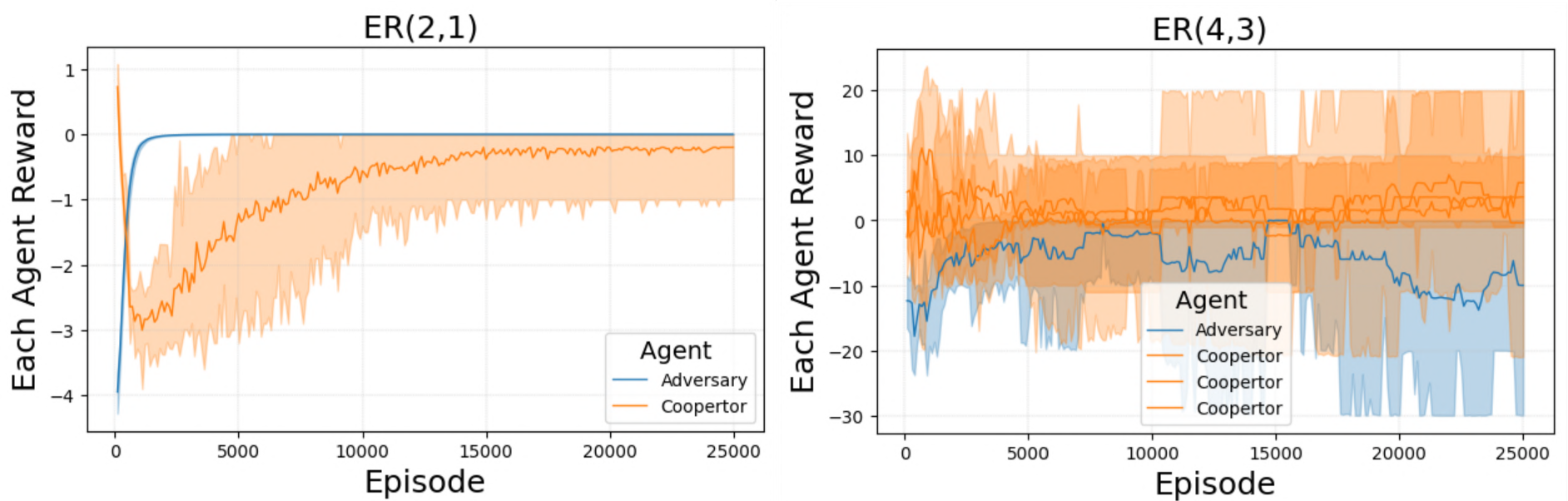}
\includegraphics[width=0.4\textwidth]{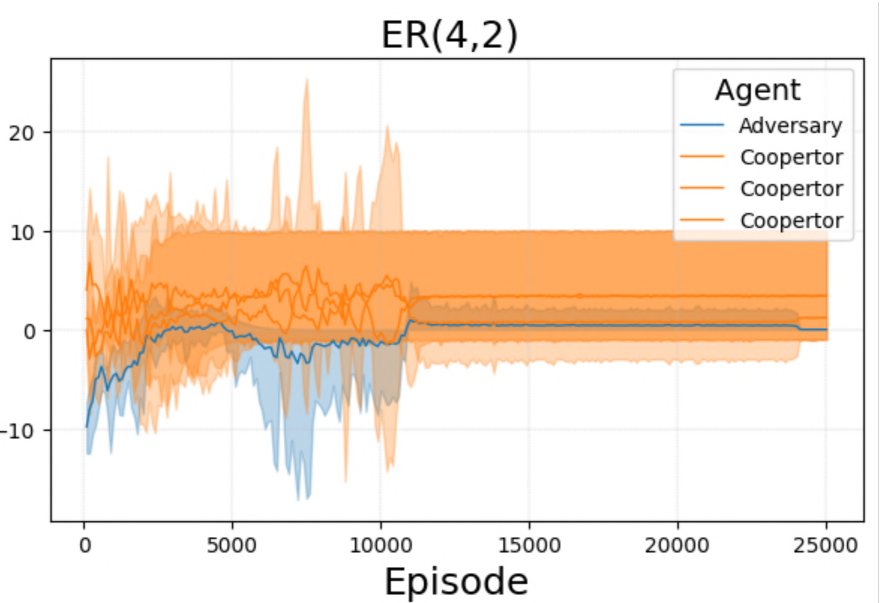}
\includegraphics[width=0.4\textwidth]{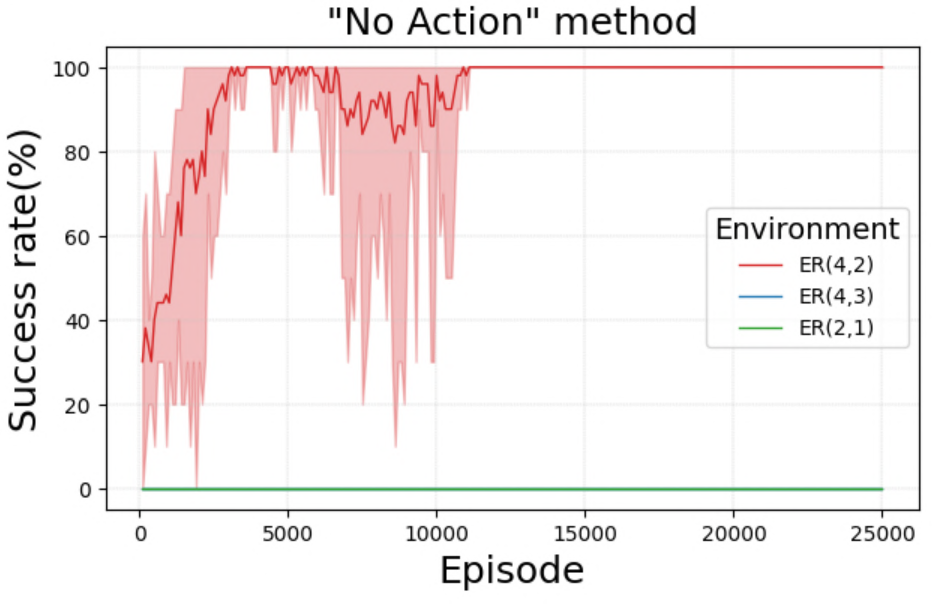}
\caption{Impact of the ``Bypass Policy'' method in ER. Cooperators send negative incentives to coax the adversary into acting. In ER(4,2), redundancy allows others to finish despite the adversary; in ER(2,1) and ER(4,3), lack of redundancy causes failure.}
\label{noaction}
\end{figure*}

\begin{figure}[!tbp]
\centering
\includegraphics[width=0.8\textwidth]{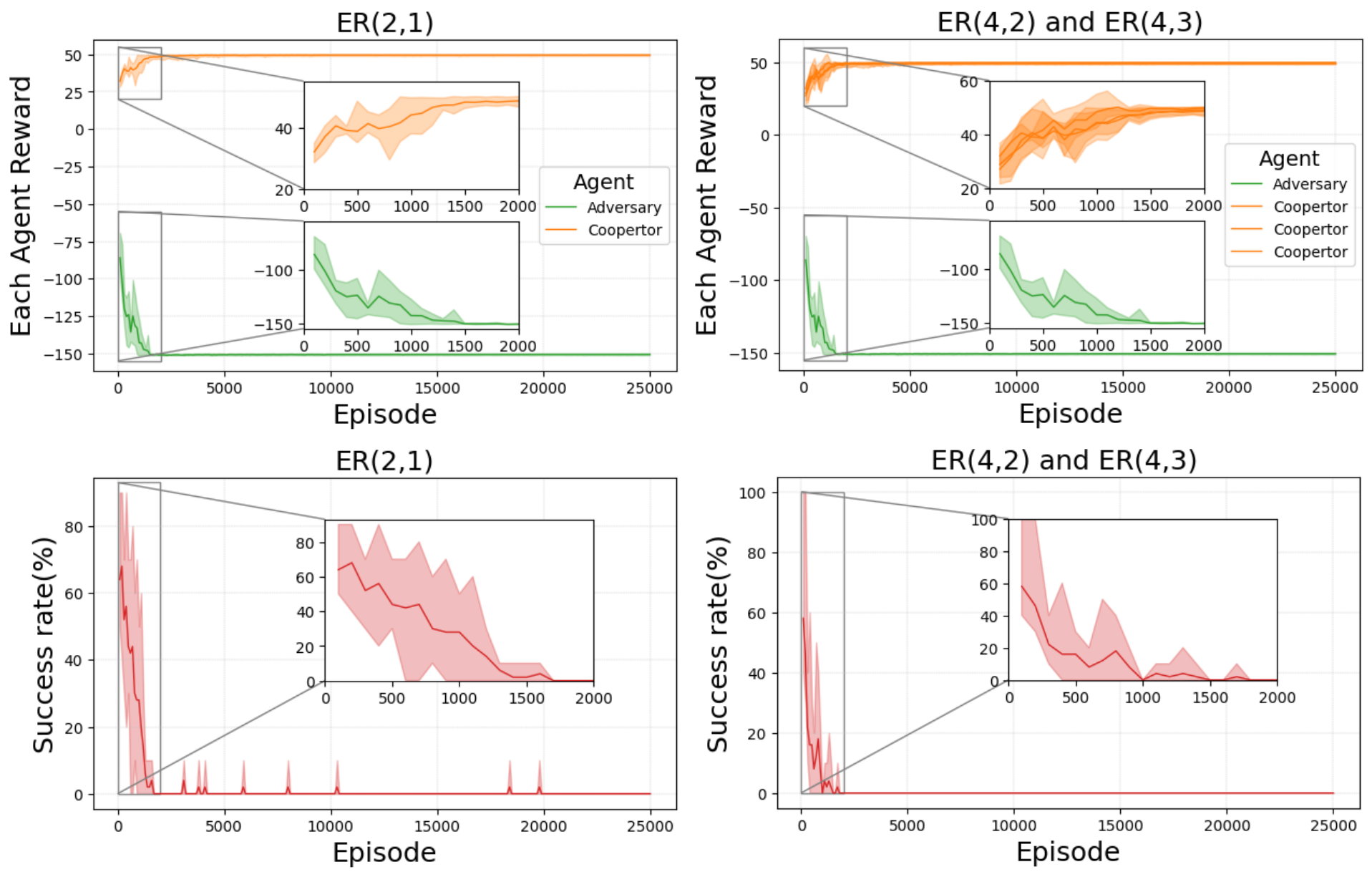}
\caption{Fake incentives destroy cooperation, producing early but unstable successes during exploration and a near-zero final success rate.}
\label{bignum}
\end{figure}

\begin{figure}[!tbp]
\centering
\includegraphics[width=0.8\textwidth]{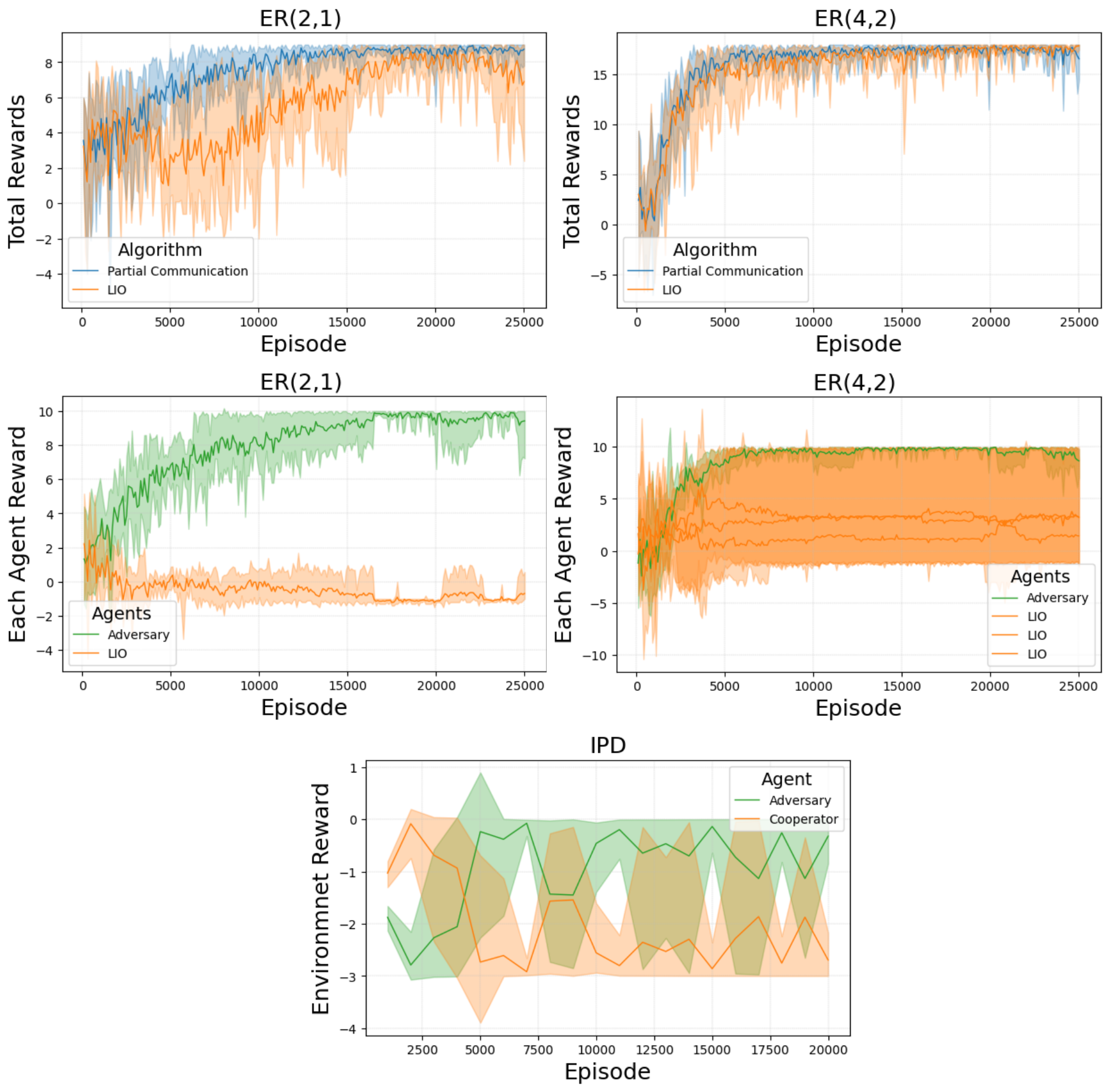}
\caption{Partial communication shortens convergence in ER. Bottom panels show the adversary’s reward dominating under this manipulation.}
\label{convergance}
\end{figure}

\subsection{Experimental Results} 

To demonstrate the effectiveness of the manipulation methods discussed in the previous section for multi-robot reinforcement learning in social dilemmas, we present the results of implementing these methods in both environments in this section. Specifically, we will discuss the results of each manipulation method used in the experiments. By evaluating the performance of our approach compared to the baseline LIO~\cite{10.5555/3495724.3496999}, we can exhibit the benefits of our proposed manipulation techniques.

\subsubsection{Incentive Reward Manipulation}
In this section, we discuss the methods used for incentive manipulation, namely partial communication and fake incentive rewards, which exploit the incentive channel between agents to achieve their objectives. 

The partial communication method controls the direction of the incentive function and thus maximizes the adversarial agent's extrinsic reward. As demonstrated in Figure~\ref{convergance}, the adversarial agent consistently achieves greater reward values than other benign agents in ER(2,1), ER(4,2), and IPD. Although the adversarial agent's goal is to maximize its own reward function, our approach enables the entire team of agents to solve the task in a shorter period of time. As shown in the results of ER(2,1) and ER(4,2) in Figure~\ref{convergance}, our approach reaches the global optimum reward faster than LIO in 37.5\% for ER(2,1) and 25\% in ER(4,2), as mentioned in Table~\ref{table:conv}.

The objective of reaching the optimal point in the IPD environment differs from that in the ER environment. In IPD, the global optimal reward is achieved when all agents reach their individual maximum reward functions, while in ER, some agents must gain less reward to assist others in reaching their maximum reward and completing the task. Our proposed method enabled an agent to reach the maximum reward in the IPD environment. However, an adversarial agent may only maximize its own extrinsic reward and cannot assist in achieving the total optimal reward for both agents. Therefore, utilizing our method with an adversarial agent may not result in the same outcomes.

In the Fake Incentive Reward manipulation method, the adversarial agent sends out malicious rewards to other agents using the incentive channel. In the Escape Room environment, we set the value of $C^{Adv}$ to 50 in Eq.\ref{fake_reward}. The reason for choosing 50 as the fake reward is that it is greater than the maximum reward that agents can receive for taking any action in the environment.
As shown in the results presented in Figure~\ref{bignum}, the agents explore the environment for 1800 episodes, which is why they achieve some success rate during this period.

\begin{table}[t]
\centering
\caption{Convergence time (episode$\times 10^3$)}
\label{table:conv}
\begin{tabular}{|c|c|c|}
\hline
\multirow{2}{*}{ Method } & \multicolumn{2}{c|}{ ER } \\
\cline{2-3} & (4,2) & (2,1) \\
\hline LIO~\cite{10.5555/3495724.3496999} & $14$ & $20$ \\
Partial Communication & $10.5$ & $12.5$ \\
\hline
\end{tabular}
\end{table}

\subsubsection{Policy Manipulation}
In the environments discussed previously, we applied the policy manipulation method, which enables the adversarial agent to manipulate the environment by altering the action caused by its policy. The following section presents the results of both the ``Bypass Policy'' and ``Reverse Policy'' methods. 

We evaluated the ``Bypass Policy'' method as the first approach to policy manipulation in the ER environment. In this method, the agent bypasses the policy, and as a result, it does not take any action and remains in its initial state. However, since social dilemmas require the cooperation and actions of all agents to reach the optimal result, this method hinders the achievement of the optimal result. The results presented in Figure~\ref{noaction} show that for ER(4,3) and ER(2,1), the agents failed to reach the optimal reward, and the success rate for those environments is zero.

\begin{figure}[!tbp]
\centering
\includegraphics[width=1.0\textwidth]{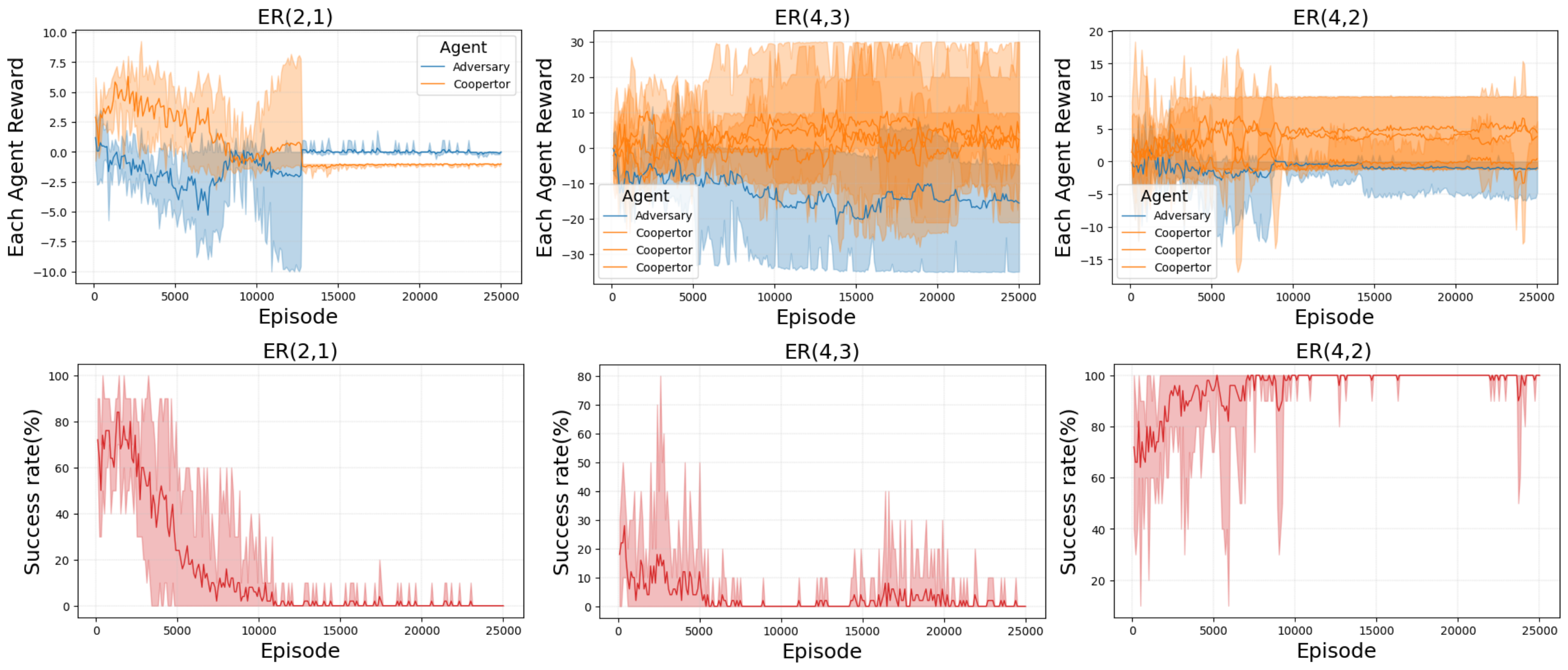}
\caption{Reverse Policy: minimizing the adversary’s own reward depresses incentives and shrinks team success, except when redundancy allows others to complete the task.}
\label{a2d}
\end{figure}

The second method of action manipulation is the ``Reverse Policy'' method, where the adversarial agent uses a different policy than other benign agents. The adversarial agent is minimizing the collective reward by changing its policy from gradient ascent to gradient descent. The adversarial agent takes actions with minimum reward and incentivizes other agents to give it the minimum incentive reward possible. The effectiveness of this method is shown in Figure~\ref{a2d}, where the success rate of task completion drops to zero.

One important observation is that environments can be classified based on whether all agents are required to collaborate to accomplish the task or whether some agents can independently contribute to achieving the goal. In this regard, we introduce the redundancy term in environments to analyze the effect of adversarial manipulation on the task completion process. When the adversarial agent abandons the main task, other agents still try to maximize their rewards. In redundant environments, other agents will ignore the adversarial agent's action and continue to move toward finishing the task. Yet, in non-redundant environments, all agents require the activity of the adversarial agent to finish the task, and they will eventually give up. In the case of our experiments, the environments of ER(4,3) and ER(2,1) are non-redundant, making the policy manipulation method successful in reducing the success rate and preventing the attainment of the total optimal reward. However, in the environment of ER(4,2), where two agents are required to pull the lever and only one agent is needed to reach the door, the adversarial agent's abandonment of other agents does not affect the final task completion process. As shown in Figure~\ref{a2d} and Figure~\ref{noaction}, agents still reach the total optimal global reward.

\subsubsection{Adaptive Multi-Objective Optimization}
\label{sec:mo-ablation}

\begin{figure*}[!tbp]
\centering 
\includegraphics[width=\linewidth]{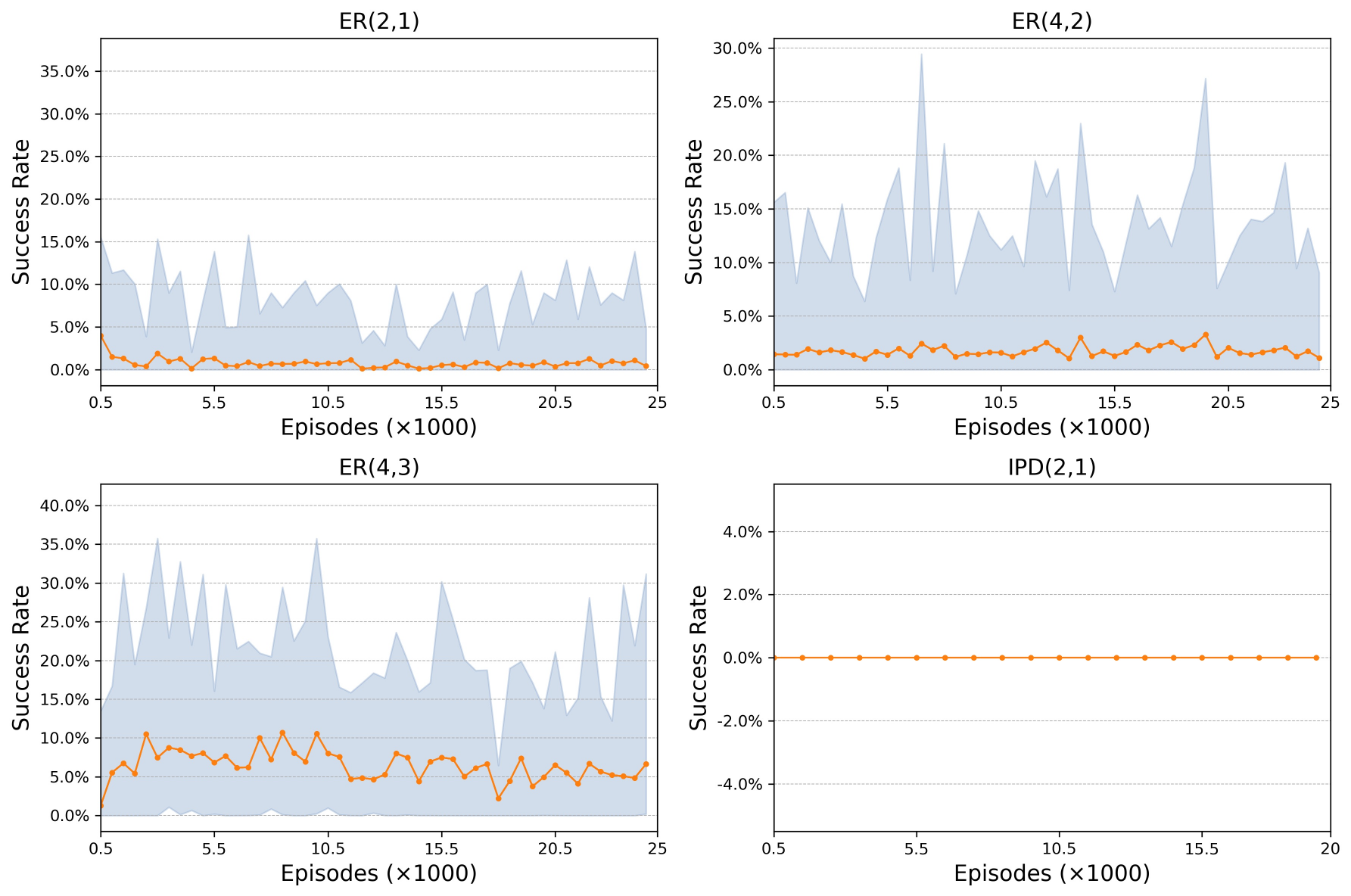}
\caption{Agent success rates across scenarios (ER and IPD) for multi-objective manipulation.   }
\label{fig:success_rates} 
\end{figure*}

\begin{figure*}[!tbp]
  \centering
  \includegraphics[width=\linewidth]{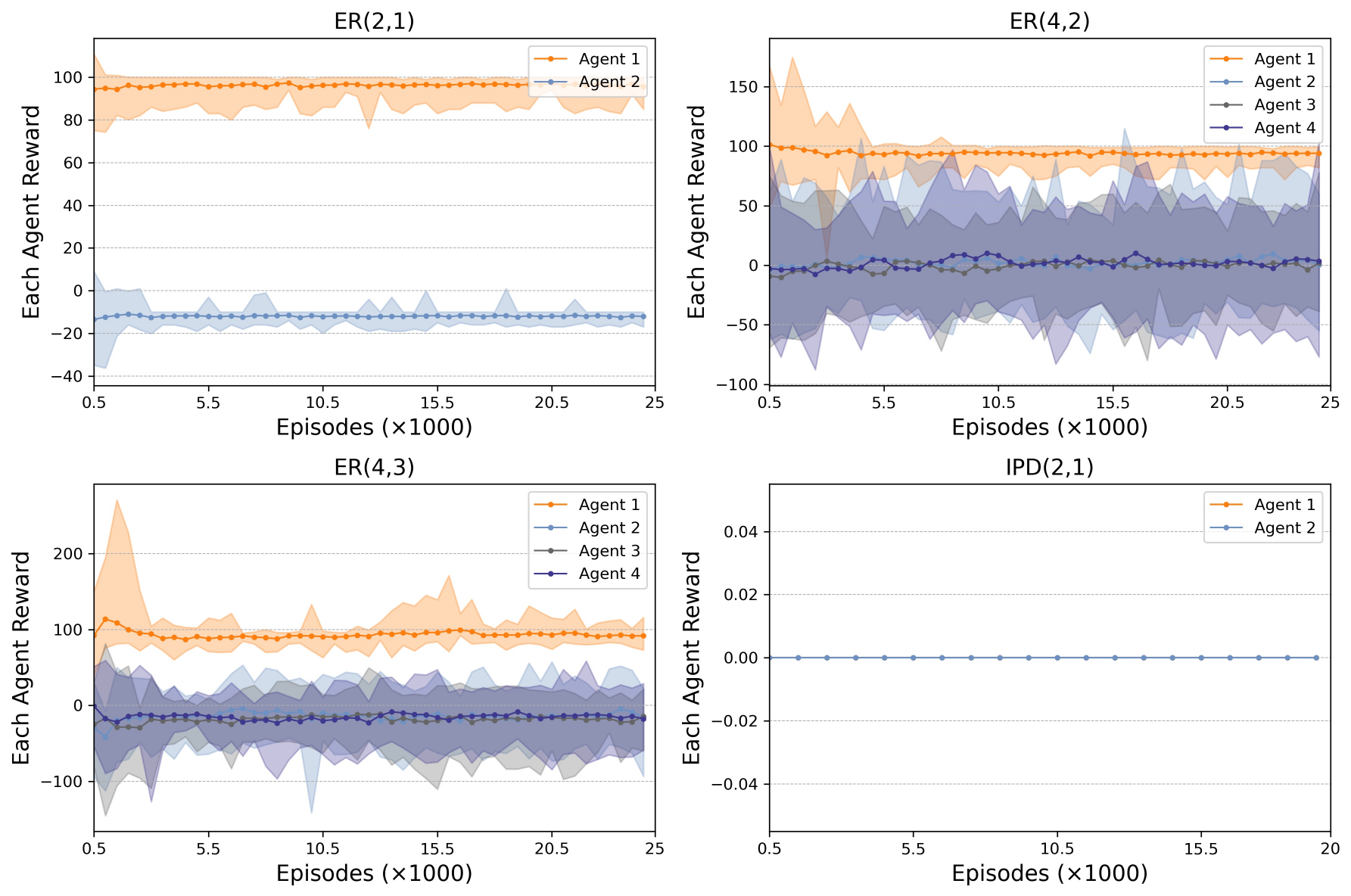}
  \caption{Per-agent rewards across scenarios (ER and IPD) for multi-objective manipulation.}
  \label{fig:agent_rewards}
\end{figure*}
 
Across the evaluated benchmarks, our proposed ADMO attack consistently demonstrates effectiveness in both ER and IPD settings. From the success-rate traces (Fig.~\ref{fig:success_rates}), we observe that ER variants show non-trivial increases for targeted actors under MO. Specifically, the target actor exhibits much larger means and wider upper tails in all ER panels, whereas the IPD(2,1) panel remains essentially flat, suggesting that the ADMO attack is more effective in a single strategy-based attacking method. 

Furthermore, we analyze per-agent rewards to in-depth showcase the effectiveness of our proposed ADMO attacking method. As demonstrated in  Fig.~\ref{fig:agent_rewards}, in ER scenarios, the adversarial agent's reward curve is noticeably elevated relative to peers, while other agents' rewards decline or remain more variable. On the other hand, in the IPD scenario, all agents' reward traces are near zero, consistent with the success rate results. 

\begin{figure*}[!tbp]
  \centering
  \includegraphics[width=\linewidth]{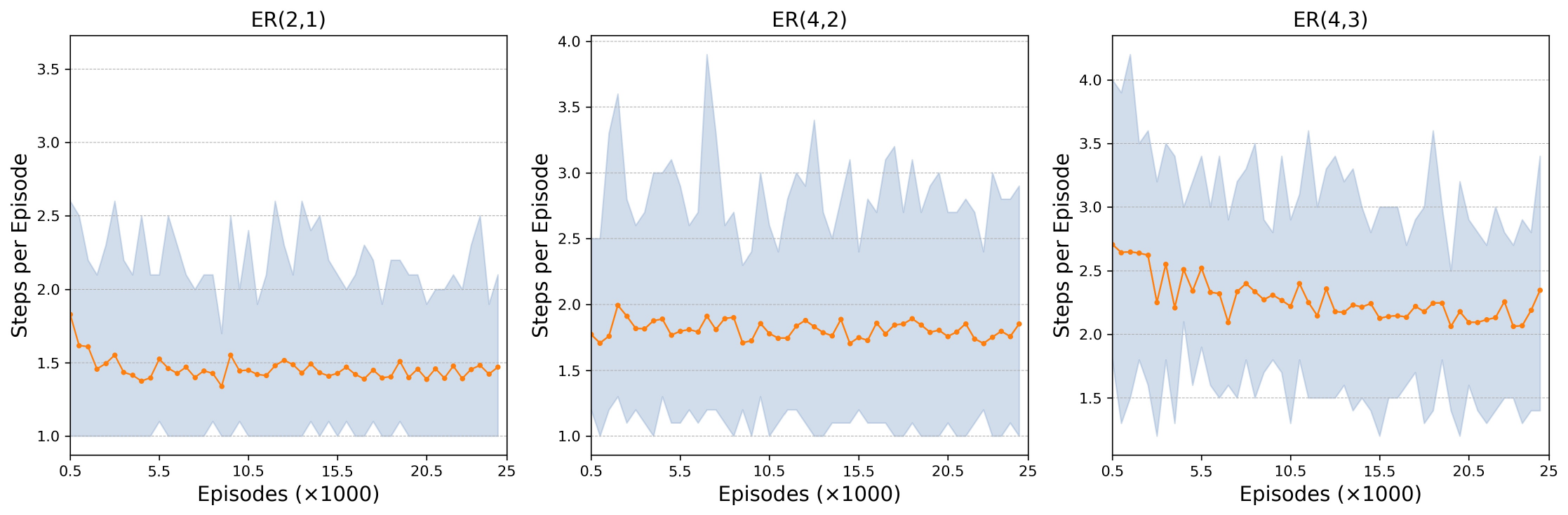}
  \caption{Average steps per episode across ER configurations for multi-objective manipulation.}
  \label{fig:steps_per_eps}
\end{figure*}

Finally, we also provide steps-per-episode diagnostics in Fig.~\ref{fig:steps_per_eps}. This metric is only provided in three ER variants. The higher steps-per-episode usually indicate a lower attacking success rate, and the lowest valid steps-per-episode value is 1.0. The experimental results show that MO runs induce modest changes in episode dynamics for ER (slightly longer and more variable episode lengths in some ER configurations), which aligns with the controller inserting additional incentive/policy computations and occasionally altering interaction patterns; again, IPD exhibits little change. Taken together, these results indicate that the proposed ADMO attack is effective in manipulating outcomes in multi-agent ER benchmarks (increasing adversary returns and success measures, and altering interaction dynamics) while having a limited impact in the simple IPD configuration tested, a useful characterization for both attack capability and domain sensitivity.

\section{Case Study on Robotic Simulation}
\begin{figure}[!tbp]
\centering
\includegraphics[width=1.0\textwidth]{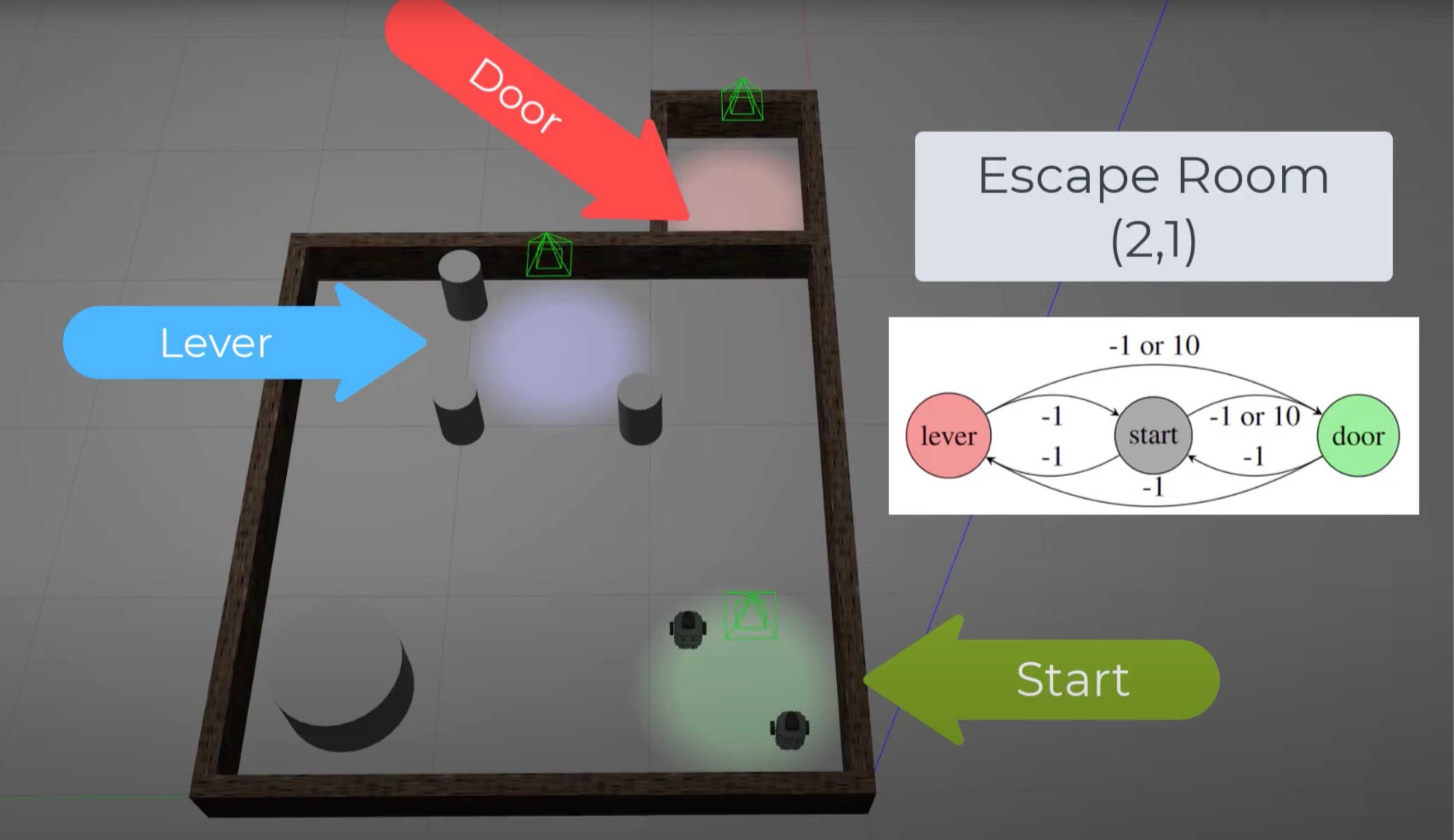}
\caption{Robotic Simulation of \Approach{} in Gazebo sumulator.}
\label{fig:robotic_simulation}
\end{figure}

To ground the discussion before moving to hardware, we begin with a controlled simulation study. We evaluate our algorithm in the open-source \emph{Gazebo} simulator, as shown in Figure~\ref{fig:robotic_simulation}, which provides high-fidelity physics, rich sensor emulation, and clean integration with ROS~2 for control and logging.\footnote{\url{https://gazebosim.org/docs/latest/tutorials/}} Simulated experiments enable rapid, reproducible evaluation across varied scenes, letting us sweep environment parameters and ablate components with minimal confounds prior to real-world trials. For intuition and transparency, we also provide a short video of the setup and results.\footnote{\url{https://youtu.be/rMqzAkxx5FY}} 

In the video clip, we showcase the Escape Room ER$(N,M)$ tasks using simulated TurtleBot3 platforms driven by ROS~2 nodes. The world encodes the canonical ER mechanics used in our study: the exit door yields a $+10$ extrinsic reward when crossed, while movement incurs a $-1$ step cost; the door opens only when $M$ teammates concurrently pull a lever, creating a genuine cooperation bottleneck. This design ensures that progress requires forming coalitions rather than exploiting single-agent shortcuts. We randomize map layouts and initial robot poses and inject modest observation/actuation noise to capture variability typical of real deployments, while keeping the underlying reward structure fixed to enable cross-run comparability.

Building on this setup, we next toggle between the two manipulation families that define \Approach{}: \emph{incentive manipulation} and \emph{policy manipulation}. For incentives, we study both partial communication (limited-message channels) and the ``fake-incentive'' condition with $C^{\text{Adv}}{=}50$ configured to dominate recipients’ returns. For policies, we consider a bypass/no-op policy (inducing idling) and a reverse/gradient-descent policy (systematically pushing behavior away from the team optimum). These conditions let us disentangle whether failures arise from distorted payoffs versus corrupted control, and how those pressures interact with coalition thresholds.

The resulting trajectories visually highlight the redundancy effect: ER$(4,2)$ remains solvable even when one agent idles because the remaining robots can still form a feasible coalition; by contrast, ER$(2,1)$ and ER$(4,3)$ fail under bypass due to tighter thresholds that leave no slack for errors or free riders. Moving from qualitative to quantitative trends, we observe (i) faster convergence under partial communication. A detailed analysis of the learning trajectories reveals that when an agent acts "self-centered" by ignoring incoming incentives, it effectively anchors the environment's non-stationarity for the remaining agents. In standard LIO, reciprocal incentive-giving can create early "reciprocity traps"—where agents continuously adjust their behavior to appease each other's shifting incentive functions, delaying convergence to the actual task objective. By freezing its response to incentives, the partial-communication agent provides a stable, predictable gradient target. This stabilization of gradient updates for the receiving agents reduces exploratory thrashing and forces the remaining agents to adapt around a fixed strategy, ultimately accelerating the team's discovery of the global optimum. We observed this effect consistently across 10 random seeds in configurations with sufficient redundancy, confirming it is rooted in the learning dynamics rather than a narrow artifact. Second, we observe (ii) success-rate collapse under fake incentives and reverse policies, where manipulated objectives or gradients steer the population toward locally consistent but globally dysfunctional equilibria. These findings mirror the learning curves we report, where sample efficiency improves with limited, truthful signals yet degrades sharply when signals or policies are adversarially perturbed.

\begin{figure*}[!tbp]
  \centering
  \includegraphics[width=\linewidth]{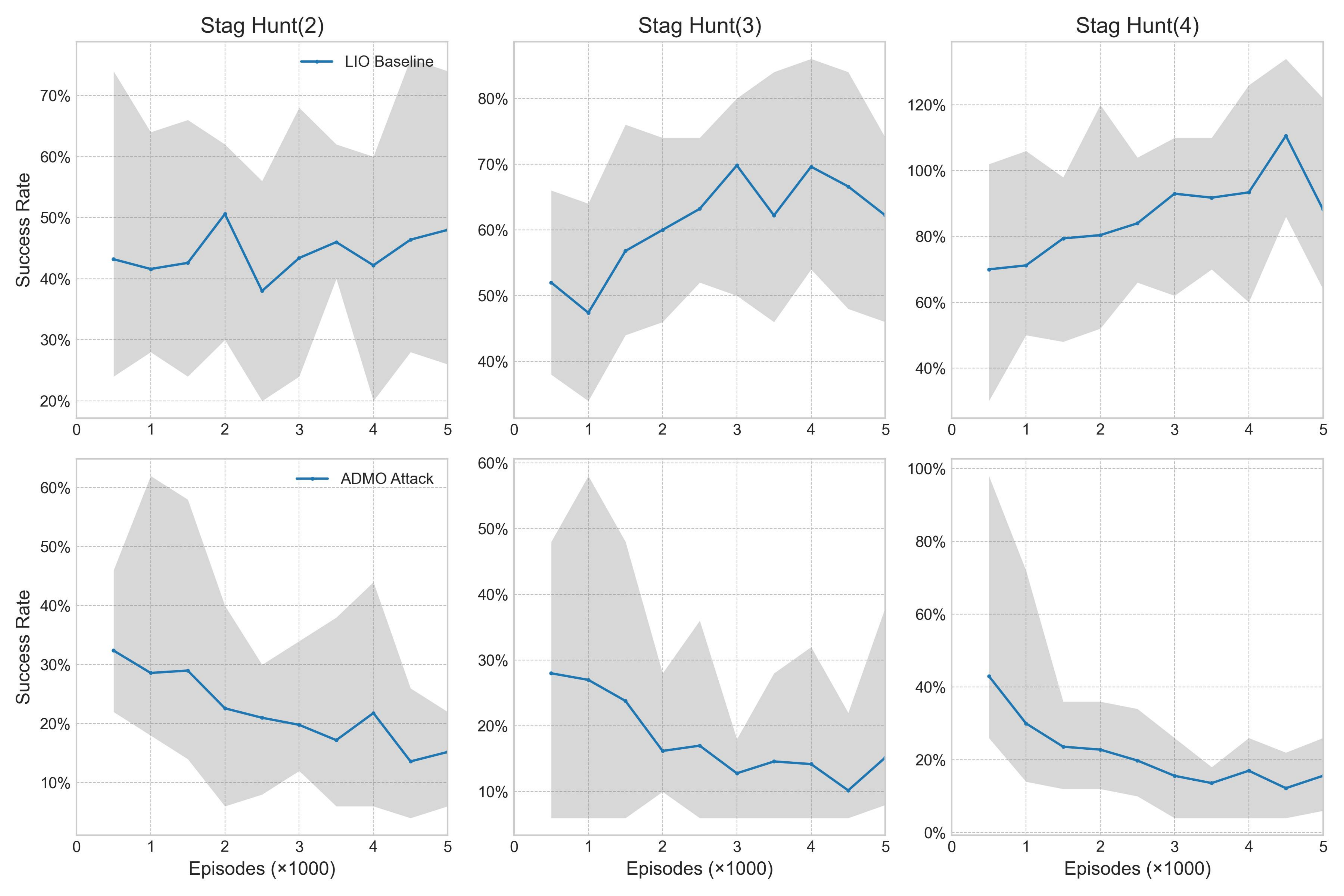}
  \caption{Agent success rates across Stag Hunt configurations ($N=2, 3, 4$) for LIO Baseline (top) and ADMO Attack (bottom).}
  \label{fig:staghunt_success}
\end{figure*}

\paragraph{Stag Hunt Success Rates} 
To rigorously address concerns regarding the benchmark complexity of ER and IPD, we evaluated \Approach{} on the more complicated Stag Hunt environment. Stag Hunt demands strict simultaneous coordination among all $N$ agents. Fig.~\ref{fig:staghunt_success} showcases the success rates for configurations of $N=2$, $3$, and $4$ agents over 5,000 episodes, averaged across 10 independent trials with varying random seeds. We define the success rate as the normalized environmental reward derived primarily from successfully capturing the stag (requiring the presence of all $N$ agents concurrently). In the top row, the LIO Baseline agents successfully learn to build trust and coordinate, leading to a gradual increase in the team success rate over time. In stark contrast, the bottom row demonstrates the devastating impact of the ADMO Attack. By actively shifting its environmental objective to penalize cooperation (i.e., hunting the hare instead of the stag), the manipulated agent completely fractures the fragile $N$-agent coalition. As training progresses, the ADMO attack continuously depresses the team's success rate towards zero. This result explicitly confirms that PIMbot scales to highly complex, trust-dependent continuous domains, maintaining the capacity to actively disrupt coordination even as the required coalition size expands.

\subsection{Extended Evaluation on SOTA MARL Algorithms}
\label{sec:sota_marl_eval}
To demonstrate the generalizability and robustness of our proposed ADMO manipulation method, we extended our evaluation to a recent state-of-the-art (SOTA) Multi-Agent Reinforcement Learning framework, Reciprocators (NeurIPS 2024). This advanced baseline incorporates reciprocal reward structures to intrinsically foster cooperation among agents, presenting a more resilient target against adversarial manipulations than standard cooperative baselines.

\begin{figure*}[!tbp]
  \centering
  \includegraphics[width=\linewidth]{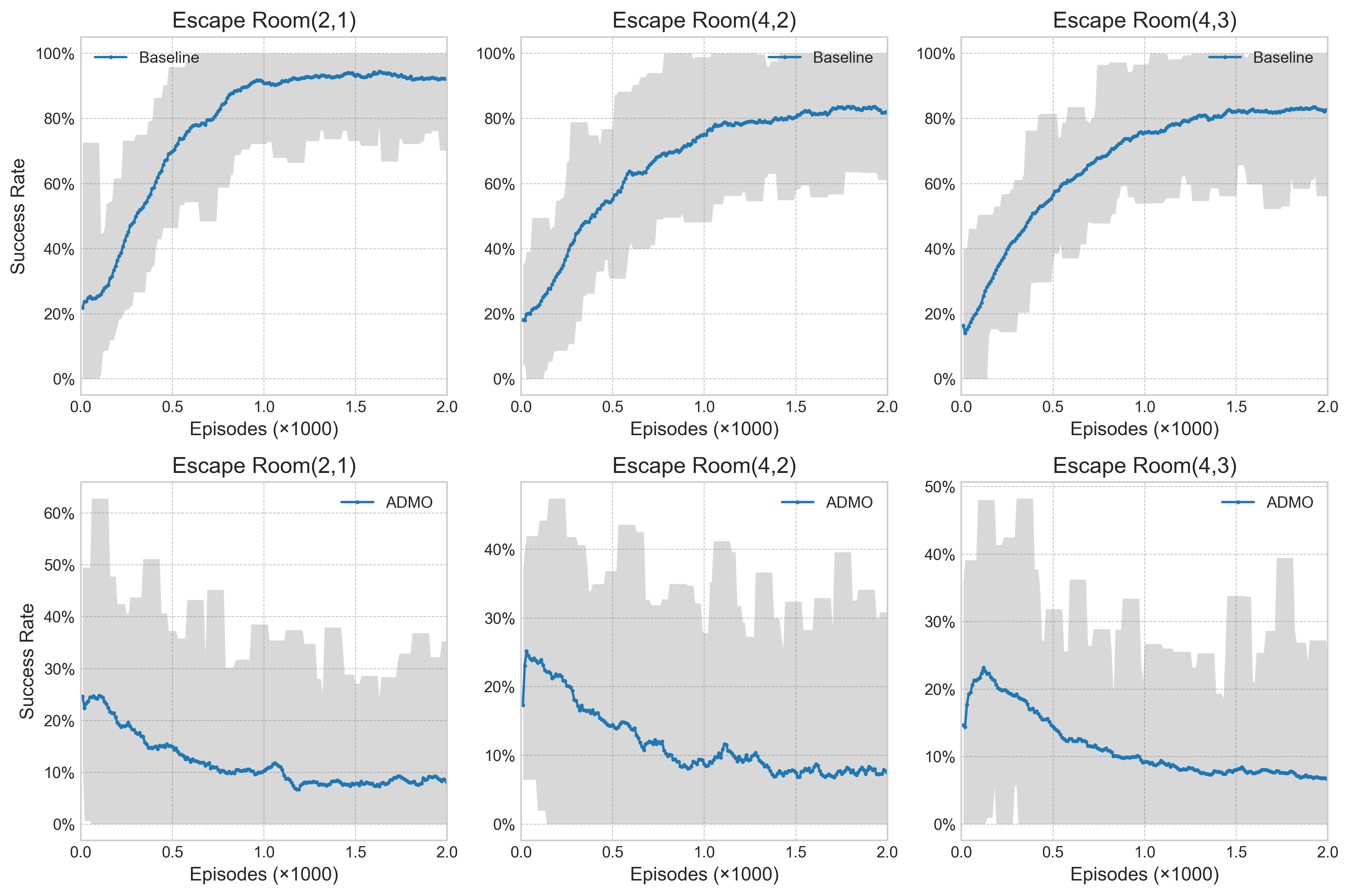}
  \caption{Agent success rates across Escape Room configurations under the SOTA Reciprocators Baseline (top) and under ADMO Attack (bottom).}
  \label{fig:er_sota_results}
\end{figure*}

We deployed our ADMO attack against the Reciprocators framework across all three environments: Escape Room (ER), Iterated Prisoner's Dilemma (IPD), and Stag Hunt. Figure~\ref{fig:er_sota_results} illustrates the success rates in the ER environment variants. While the cooperative Reciprocators baseline quickly establishes a high success rate (reaching over 80\%--90\% across ER configurations), the introduction of an ADMO adversary systematically subverts this coordination. The manipulated agent actively bypasses cooperative conventions, driving the team's success rate down to near zero in ER(4,3) and maintaining it below 30\% in ER(4,2), confirming that ADMO can effectively disrupt even sophisticated reward-sharing mechanisms.

\begin{figure*}[!tbp]
  \centering
  \includegraphics[width=\linewidth]{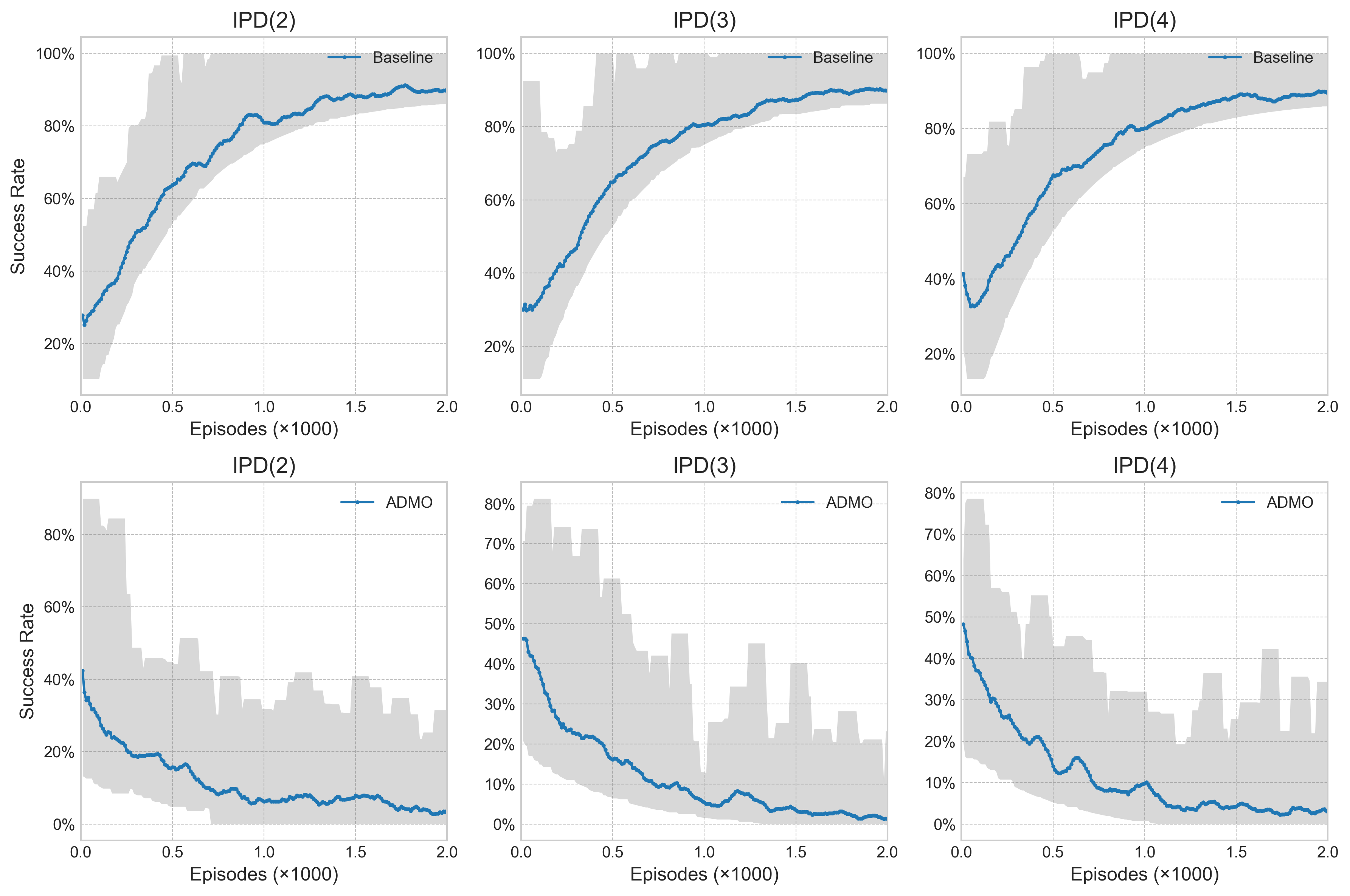}
  \caption{Agent success rates across IPD configurations under the SOTA Reciprocators Baseline (top) and under ADMO Attack (bottom).}
  \label{fig:ipd_sota_results}
\end{figure*}

In the IPD environment (Figure~\ref{fig:ipd_sota_results}), we observe a similar vulnerability. The Reciprocators algorithm achieves near-optimal cooperation (approaching 90\% success rate) in IPD(2), IPD(3), and IPD(4). However, when subjected to the ADMO attack, the targeted agent aggressively exploits the baseline's reciprocal nature, causing the success rate to plummet significantly across all configurations (dropping below 10\%), thereby forcing the team into suboptimal, defection-heavy equilibria.

\begin{figure*}[!tbp]
  \centering
  \includegraphics[width=\linewidth]{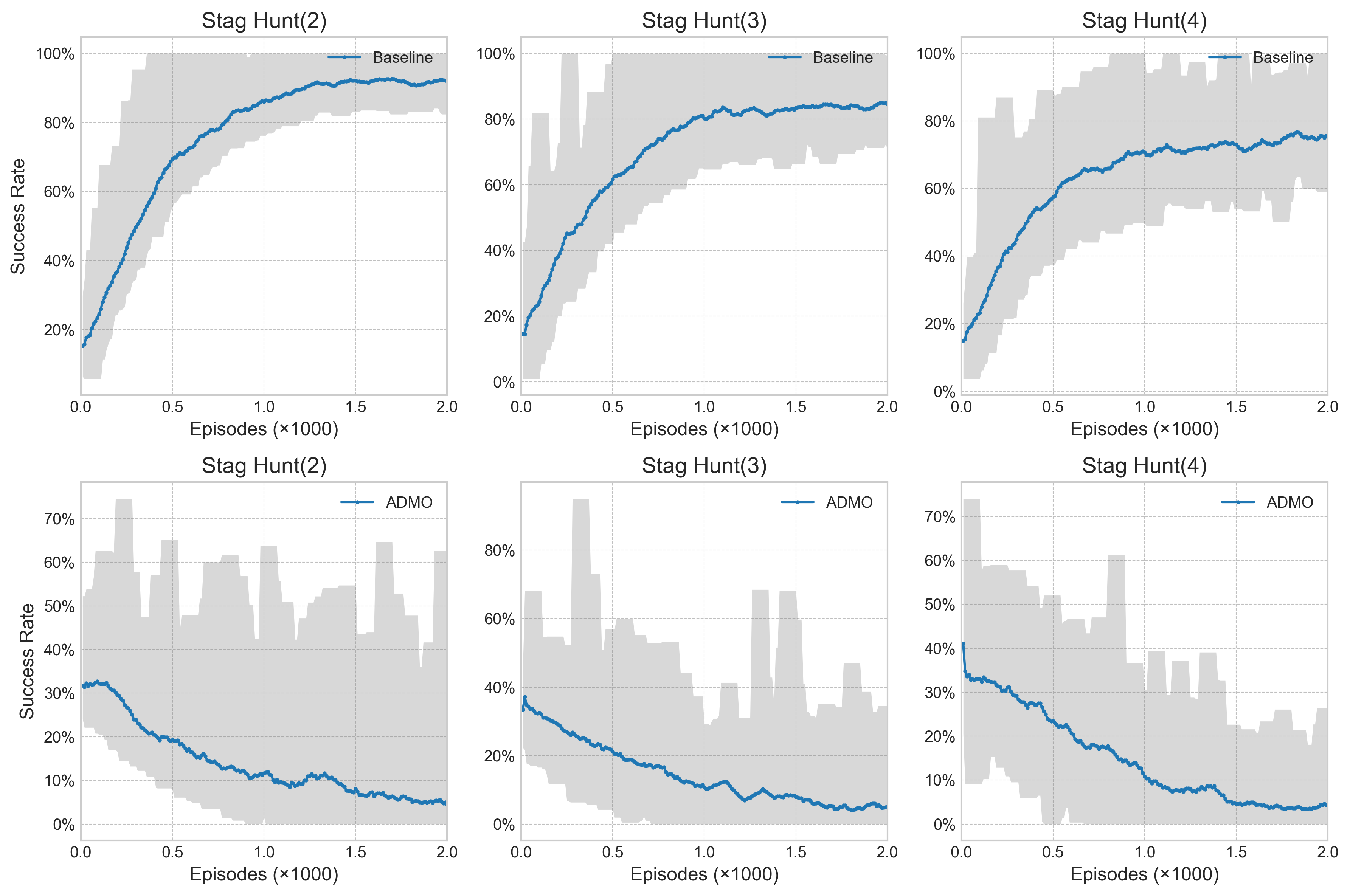}
  \caption{Agent success rates across Stag Hunt configurations under the SOTA Reciprocators Baseline (top) and under ADMO Attack (bottom).}
  \label{fig:staghunt_sota_results}
\end{figure*}

Finally, we evaluated the ADMO attack against Reciprocators in the highly complex Stag Hunt environment (Figure~\ref{fig:staghunt_sota_results}), which requires strict, simultaneous coordination. Despite the baseline's strong capabilities in forming stable coalitions (achieving up to 80\%--90\% success), the ADMO adversary reliably fractures these alliances. As shown in the results for Stag Hunt(2), Stag Hunt(3), and Stag Hunt(4), the ADMO attack causes an irreversible collapse in coordination, effectively driving the success rate to near zero. These comprehensive results explicitly validate that our manipulation framework is not merely an artifact of simpler algorithms, but a potent and scalable threat against modern, state-of-the-art MARL systems.

\section{Case Study on Real Embedded Systems}
\label{sec:jetson}

To assess deployability beyond simulation and quantify the practical overheads of manipulation at the edge, we next move to a real, GPU-enabled embedded platform. Specifically, we deploy \Approach{} on an NVIDIA Jetson Orin Nano and profile latency, utilization, power, and thermal behavior under benign and adversarial modes. As a design choice to reflect a common robotics stack, we place both the policy and incentive heads on-device while keeping the environment simulation off-device; any network round-trip time (RTT) is hidden behind action repeat to stabilize pacing. 
Jetson Orin Nano (8\,GB) is configured in 15\,W MaxN mode on \mbox{Ubuntu~20.04} with \mbox{JetPack~5.1.2}. We use Python 3.8.20 to execute the policy and incentive MLPs. CPU affinity is pinned to reduce OS jitter, and dynamic voltage and frequency scaling (DVFS) remains enabled to reflect realistic power management. Profiling is performed with \texttt{tegrastats} at 1 Hz. The receiver and giver heads match the simulation architecture (two hidden layers: 256 and 128; ReLU; softmax policy). Unless otherwise noted, batch size is $1$ (online control) and the ER tasks use an action repeat of $r{=}2$ for stable wall-clock pacing. 
Decision latency is measured from observation-ready to action-commit; throughput is decisions/s. Average power comes from the on-board INA3221 rail reading, and energy consumption is estimated by cumulative power in profiling. Peak temperature is the maximum thermal junction (tj) reading observed per run. Unless stated otherwise, initialization transients (process launch, ROS graph discovery) are excluded when aggregating steady-state statistics. This methodology enables us to separate control-loop costs from unrelated startup effects, allowing for apples-to-apples comparisons across controllers.

We run paired experiments on two benchmarks, Iterated Prisoner’s Dilemma (IPD) and Escape Room (ER), comparing our adaptive multi-objective (ADMO) attack controller to clean trials without multi-objective reweighting (w/o MO). To illustrate how overheads scale with platform architectures, we profile using two MARL frameworks: LIO~\cite{10.5555/3495724.3496999} (CPU-only) and Reciprocators (GPU-accelerated). The only difference across workloads is the controller: \emph{benign} (no manipulation) versus \emph{adversarial} (ADMO), thereby isolating the incremental cost of manipulation logic.

\begin{figure*}[!htbp]
\centering
\begin{subfigure}[t]{0.35\linewidth}
  \centering
  \includegraphics[width=\linewidth]{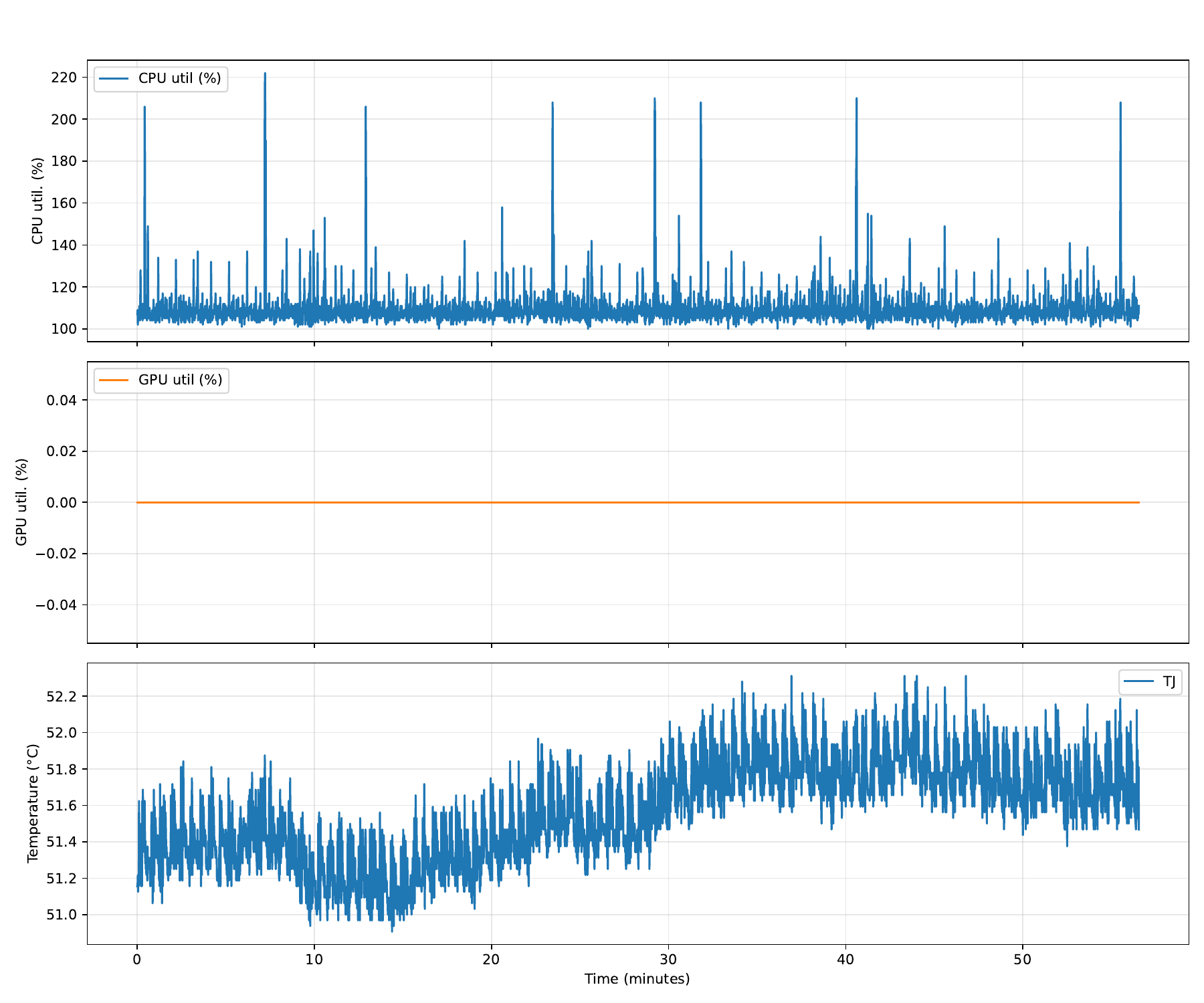}
  \caption{IPD benchmark, with our proposed Multi-Objective (MO) attack.}
  \label{fig:ipd_mo}
\end{subfigure}
\begin{subfigure}[t]{0.35\linewidth}
  \centering
  \includegraphics[width=\linewidth]{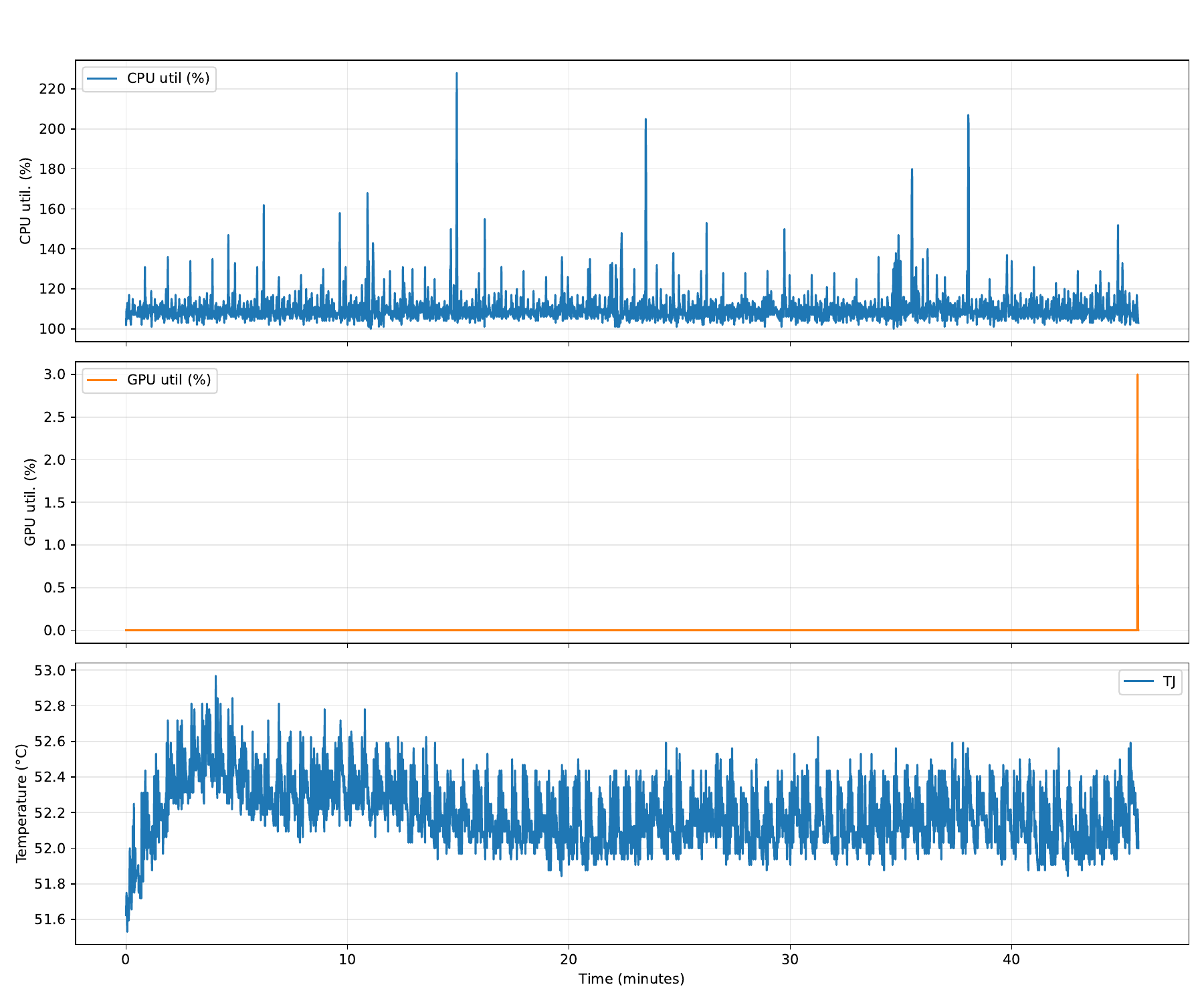}
  \caption{IPD benchmark (clean baseline). }
  \label{fig:ipd_clean}
\end{subfigure}

\vspace{4mm}

\begin{subfigure}[t]{0.35\linewidth}
  \centering
  \includegraphics[width=\linewidth]{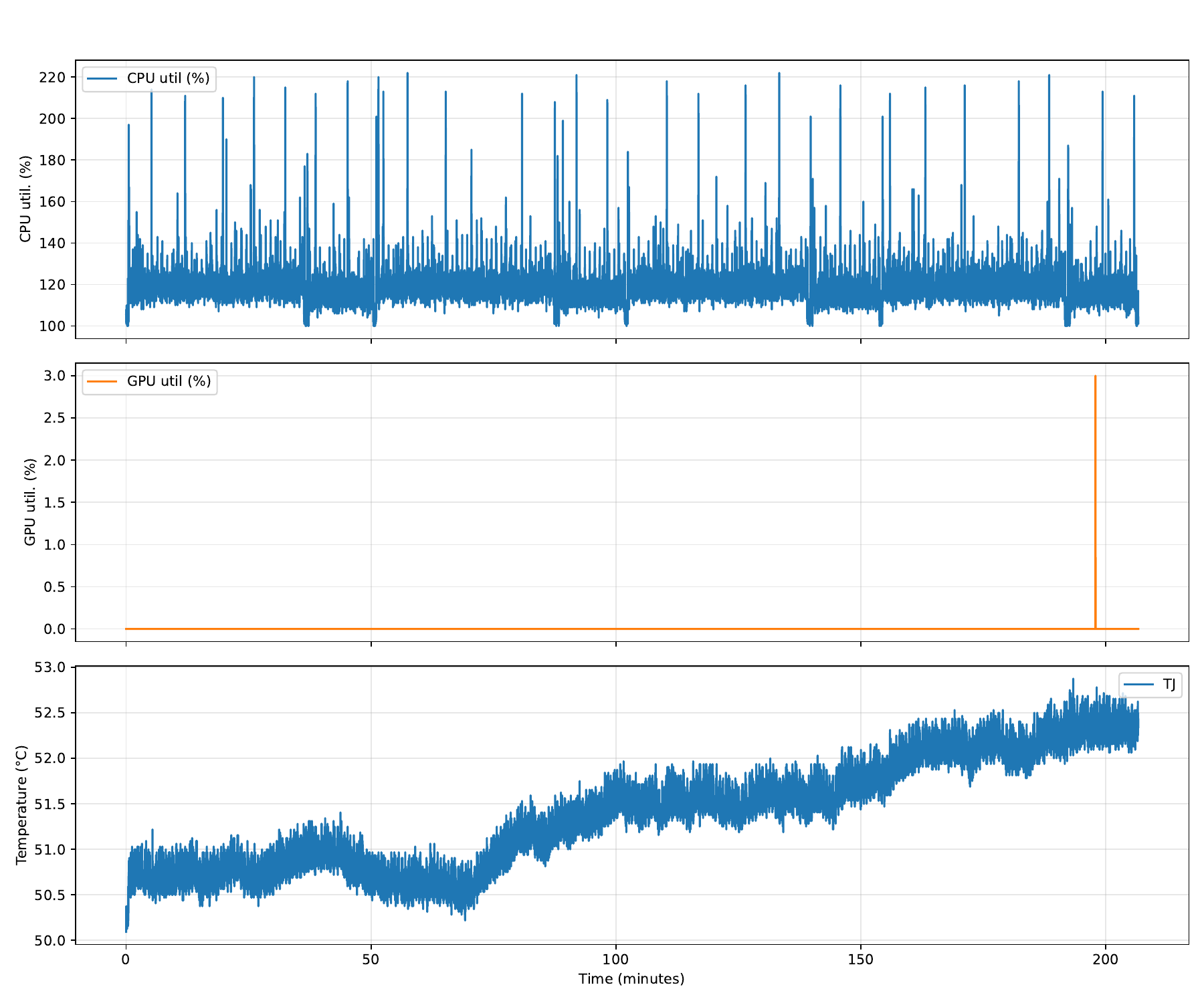}
  \caption{ER benchmark, with our proposed Multi-Objective (MO) attack. }
  \label{fig:er_mo}
\end{subfigure}
\begin{subfigure}[t]{0.35\linewidth}
  \centering
  \includegraphics[width=\linewidth]{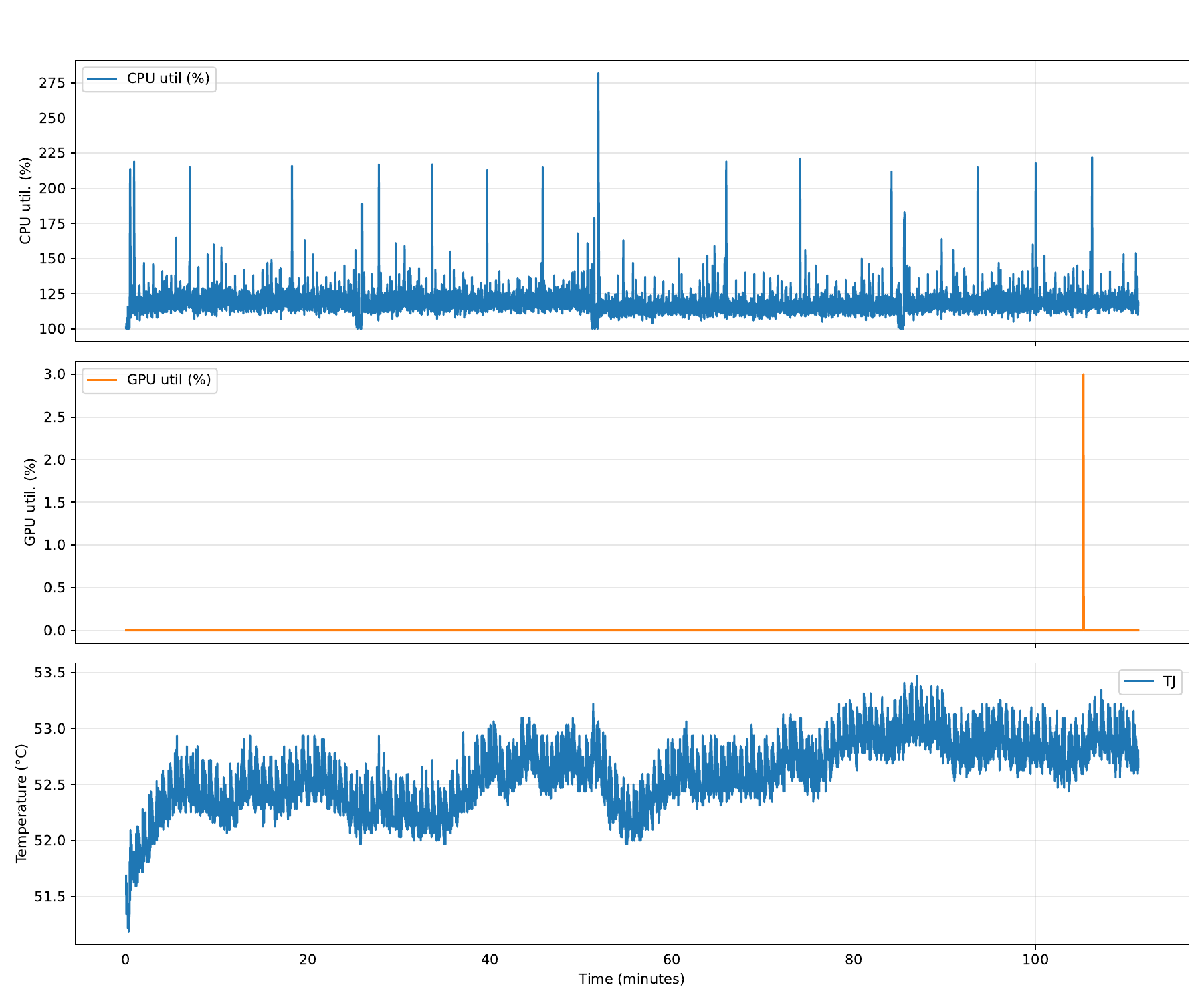}
  \caption{ER benchmark (clean baseline). }
  \label{fig:er_clean}
\end{subfigure}

\caption{System profiling traces for IPD and ER benchmarks under the LIO framework (CPU-only). Each panel contains (top) aggregated CPU utilization (percent, multiple logical cores aggregated as shown in traces), (middle) observed GPU utilization (percent), and (bottom) package/processor temperature (°C). Panels (a)/(c) run the ADMO multi-objective manipulation (``MO'') attached to the adversary; panels (b)/(d) show clean opposite trials (``w/o MO'').}
\label{fig:profiling_all}
\end{figure*}

\begin{figure*}[!htbp]
\centering
\begin{subfigure}[t]{0.35\linewidth}
  \centering
  \includegraphics[width=\linewidth]{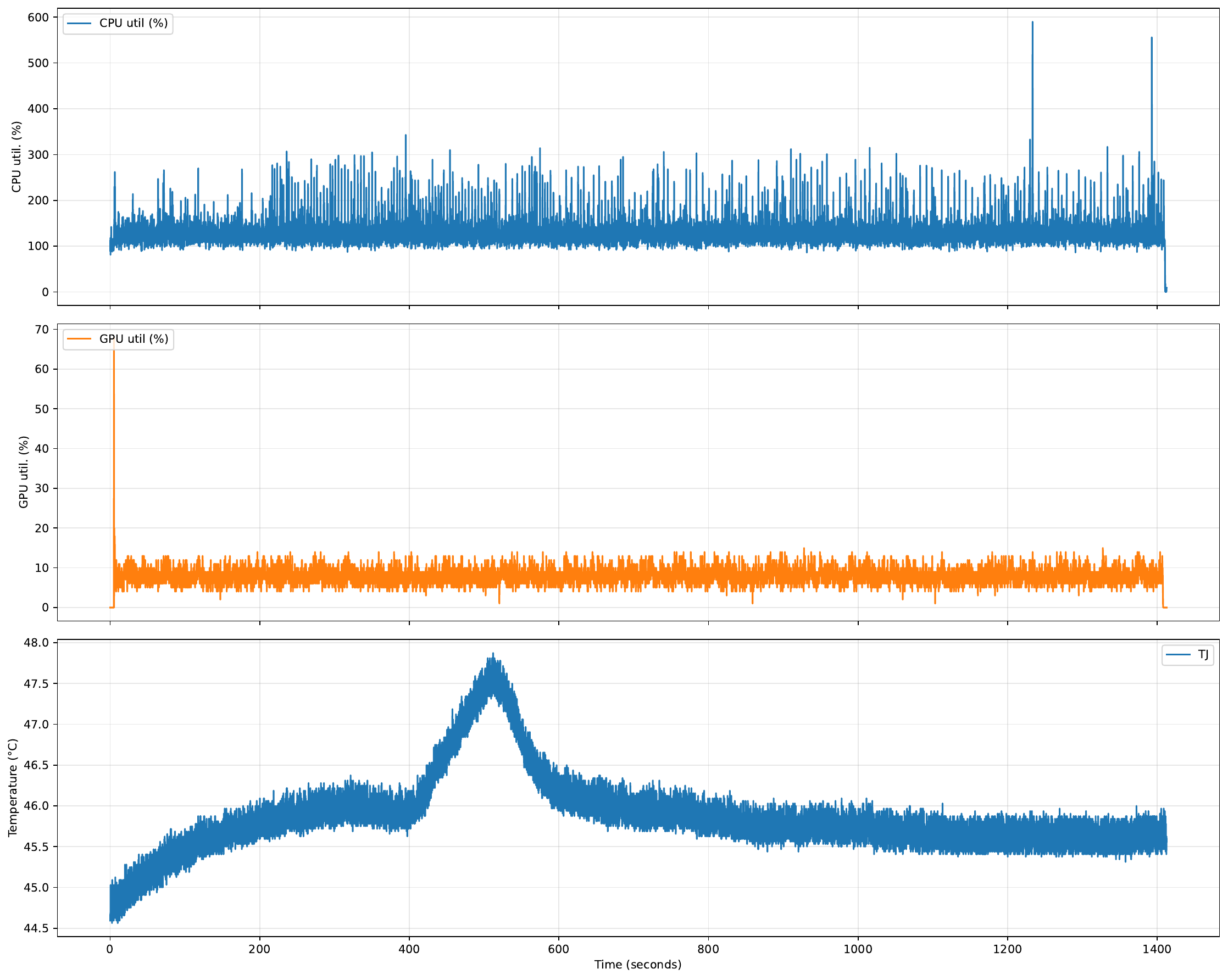}
  \caption{IPD benchmark, with our proposed Multi-Objective (MO) attack (truncated zeros).}
  \label{fig:ipd_mo_new}
\end{subfigure}
\begin{subfigure}[t]{0.35\linewidth}
  \centering
  \includegraphics[width=\linewidth]{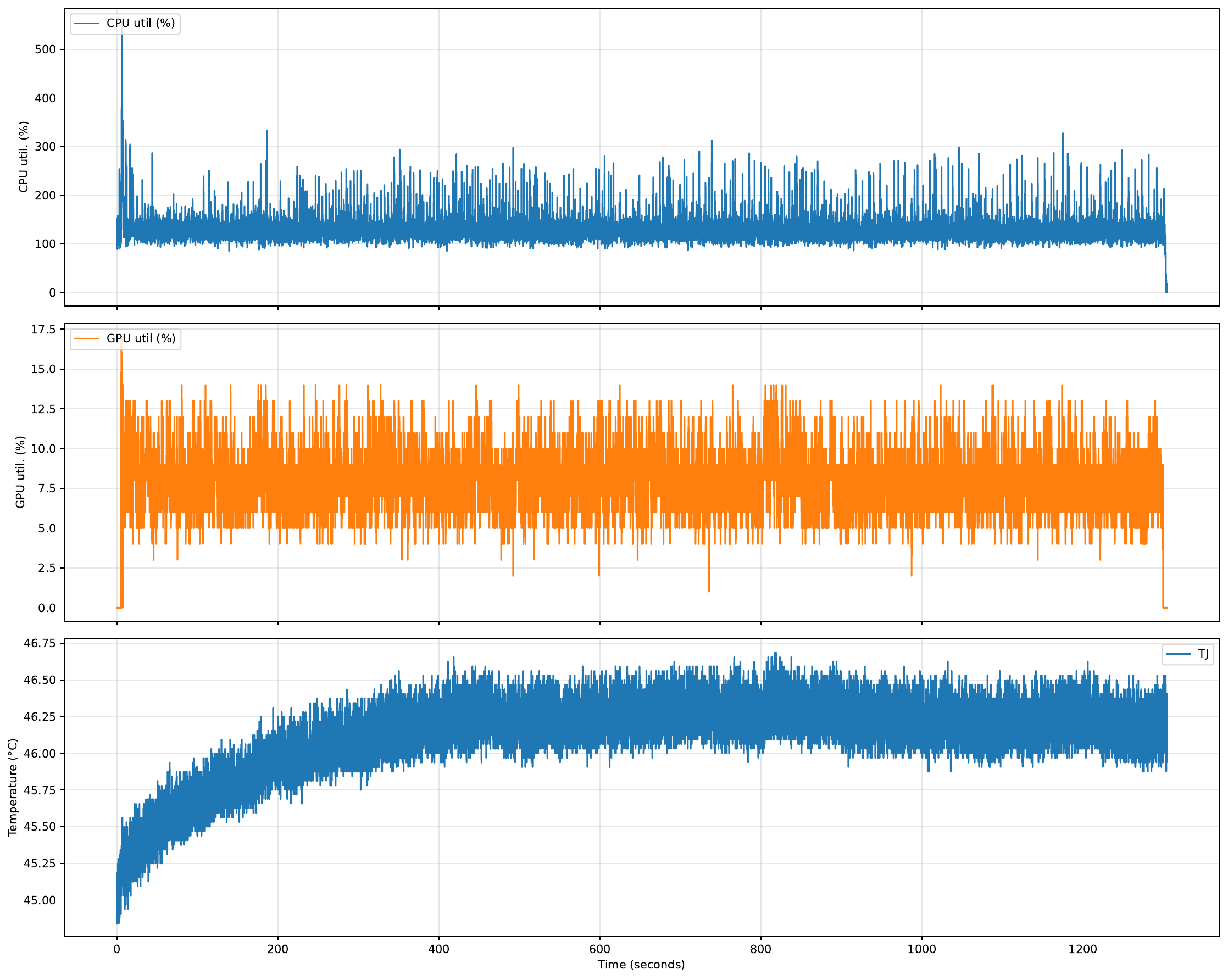}
  \caption{IPD benchmark (clean baseline, truncated zeros).}
  \label{fig:ipd_clean_new}
\end{subfigure}

\vspace{4mm}

\begin{subfigure}[t]{0.35\linewidth}
  \centering
  \includegraphics[width=\linewidth]{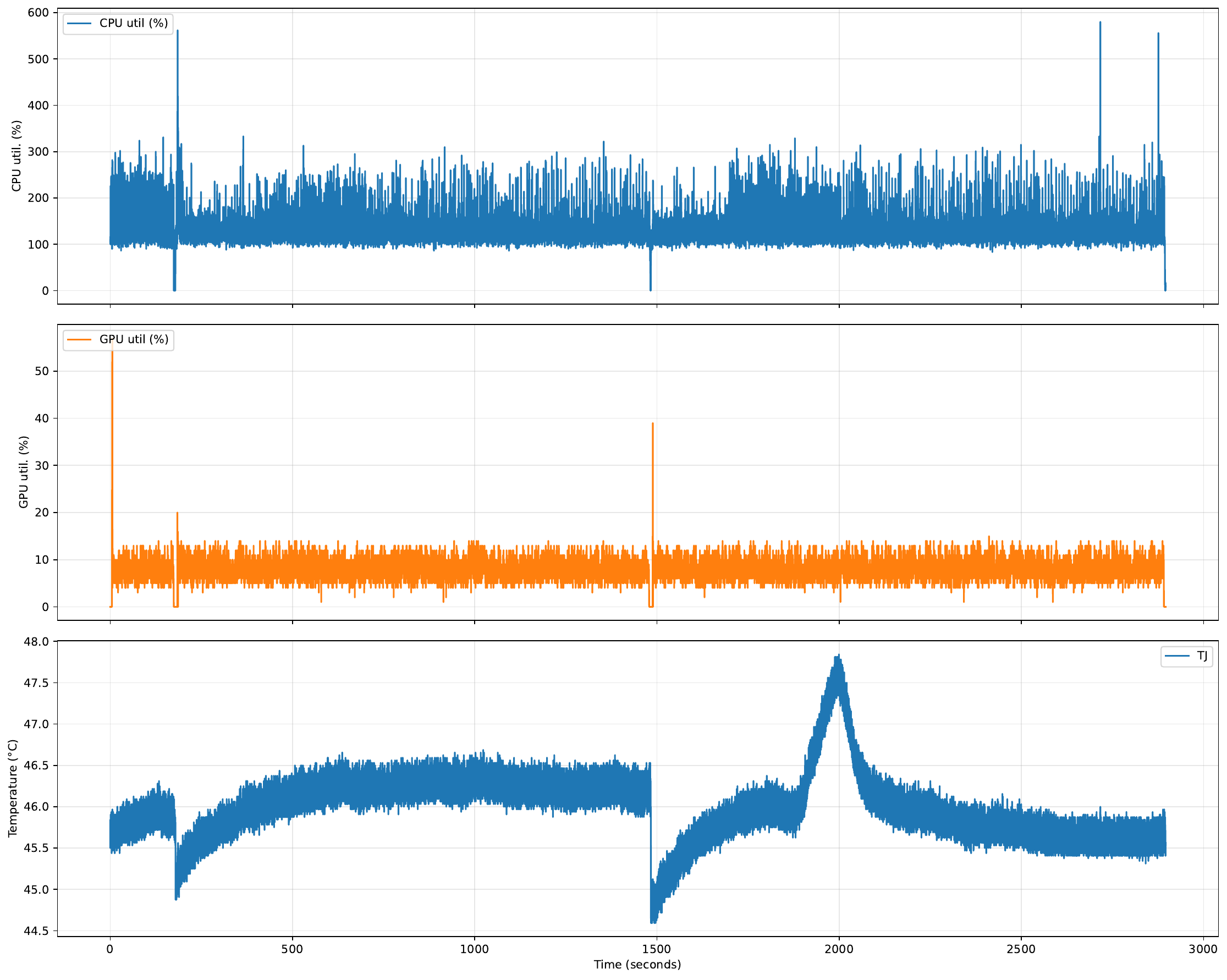}
  \caption{ER benchmark, with our proposed Multi-Objective (MO) attack (truncated zeros).}
  \label{fig:er_mo_new}
\end{subfigure}
\begin{subfigure}[t]{0.35\linewidth}
  \centering
  \includegraphics[width=\linewidth]{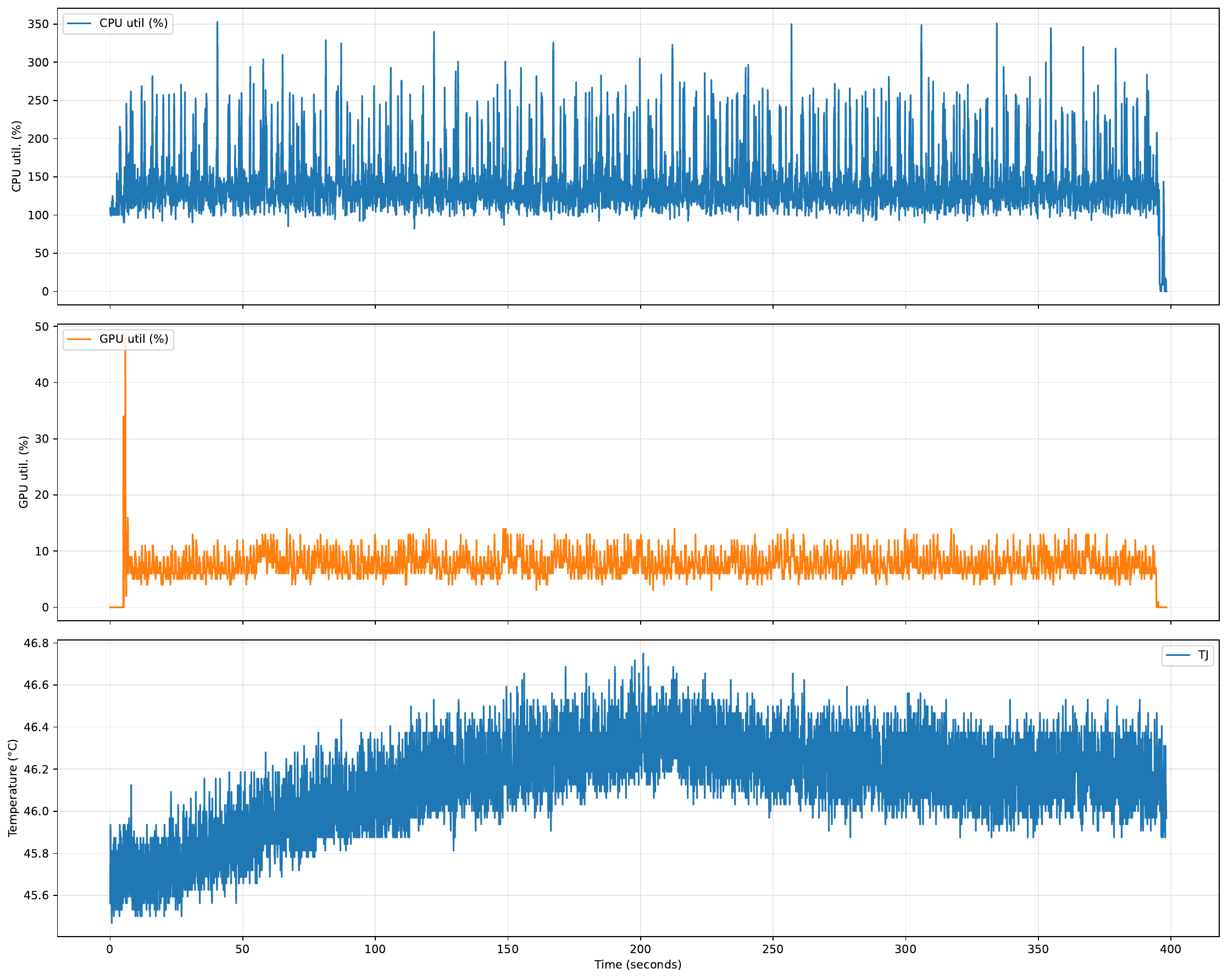}
  \caption{ER benchmark (clean baseline, truncated zeros).}
  \label{fig:er_clean_new}
\end{subfigure}

\caption{System profiling traces for IPD and ER benchmarks with truncated long zero-GPU utilization periods. Each panel contains (top) aggregated CPU utilization (percent), (middle) observed GPU utilization (percent), and (bottom) package/processor temperature (°C). Panels (a)/(c) run the ADMO multi-objective manipulation (``MO'') attached to the adversary; panels (b)/(d) show clean opposite trials (``w/o MO'').}
\label{fig:profiling_all_new}
\end{figure*}

\paragraph{System-level observations.}
The traces for the LIO framework (Fig.~\ref{fig:profiling_all}) and the Reciprocators framework (Fig.~\ref{fig:profiling_all_new}) reveal several consistent patterns across both benchmarks: (1) \textbf{CPU duty cycle.} All runs show short-lived CPU spikes, but MO-enabled traces (Fig.~\ref{fig:ipd_mo}, Fig.~\ref{fig:er_mo}, Fig.~\ref{fig:ipd_mo_new}, Fig.~\ref{fig:er_mo_new}) exhibit higher peaks and denser spike trains than their w/o-MO counterparts (Fig.~\ref{fig:ipd_clean}, Fig.~\ref{fig:er_clean}, Fig.~\ref{fig:ipd_clean_new}, Fig.~\ref{fig:er_clean_new}). This is consistent with the ADMO controller introducing additional CPU-bound work—lightweight gradient updates, small QP/KKT-style solves, and bookkeeping of exponential moving averages and signals. In effect, manipulation logic manifests as bursty CPU utilization layered atop the baseline control loop. (2) \textbf{GPU utilization.} In the LIO traces (Fig.~\ref{fig:profiling_all}), GPU activity remains negligible because LIO provides CPU-only workloads. In contrast, the Reciprocators framework (Fig.~\ref{fig:profiling_all_new}) leverages the GPU. Here, we truncate prolonged zero-utilization periods to highlight the active computation bursts. Both clean and MO-enabled trials exhibit sustained GPU utilization. Crucially, the incremental overhead of ADMO on the GPU is relatively transparent and scales naturally, an important consideration for platforms where GPU headroom is reserved for perception. (3) \textbf{Thermals and stability.} Temperature traces show a gradual upward drift over longer ER runs and modest mid-run rises in IPD. MO runs tend to produce slightly higher temperature baselines and larger variance, which aligns with the elevated and spikier CPU duty cycle under adaptive weighting. The GPU utilization in Reciprocators also contributes slightly steeper thermal increases. Nevertheless, signals remain smooth at the timescale of seconds and do not indicate prolonged, sustained saturation. (4) \textbf{Signal smoothness.} Clean (w/o-MO) traces are comparatively smoother in both CPU and temperature channels, implying that multi-objective reweighting increases short-term variability. Importantly, this variability is \emph{transient} and bounded by the controller’s update cadence.

Taken together, these on-device profiles indicate that the ADMO method imposes a modest, primarily CPU-side overhead that increases transient load and thermal variance but does not trigger sustained saturation across devices. Note that while LIO incurs primarily CPU-side overhead, the Reciprocators profiles confirm ADMO successfully operates alongside GPU-bound MARL workloads without degrading device stability. From a design standpoint, this suggests that: (i) manipulation-aware control can co-exist with perception stacks on the same SoC if GPU resources are budgeted for vision; (ii) pinning threads and preserving DVFS yields realistic power behavior while containing jitter; and (iii) the main lever for runtime cost is the frequency and complexity of the ADMO update step (e.g., batchless online updates versus minibatched solves).

\section{Discussion \& Future Works}
We explored ways to \emph{manipulate} cooperate in multi-robot RL, primarily as a stress test for adversarial robustness. Our adaptive MO controller decides online whether to lean on incentive or policy manipulation and in which adversarial mode (stealthy vs. active disruption). 
Safeguards include: (i) mandatory exploration independent of received incentives; (ii) budget caps and sanity checks on inbound incentives; and (iii) centralized audit of policies/reward flows. 
On-device, embedded constraints (latency/power/thermal) may couple to the choice of manipulation mode and weights; the Jetson study will quantify this coupling. In future work, we will consider certified defenses against fake incentives, communication authentication, and convergence analysis of Eq.~\eqref{eq:pareto_grad} under partial observability.

\section{Related Work}

\subsection{Multi-agent Reinforcement Learning}

Research on multi-agent reinforcement learning is still consolidating its foundations. Most modern systems follow centralized training with decentralized execution, where global information is leveraged by a centralized critic or trainer while policies act locally at test time, but performance remains highly sensitive to core difficulties such as non-stationarity, partial observability, exploration burden, and multi-agent credit assignment \citep{busoniu2008comprehensive,shoham2009multiagent,zhang2021multiagent,papoudakis2019dealing,chendycodeeval,li2025dr}. Within this design, cooperative value factorization decomposes a joint action value into per-agent utilities to enable decentralized execution, progressing from additive and monotonic schemes to more expressive mixers \citep{sunehag2018value,rashid2018qmix,son2019qtran,wang2021qplex}. To improve coordination under limited view, learned communication protocols and attention-based message passing have been proposed, while mean field and graph structured formulations target scalability, and opponent-aware updates shape learning in mixed settings \citep{sukhbaatar2016learning, singh2019learning, yang2018mean,foerster2018learning}. Exploration remains a bottleneck in cooperative tasks due to the large joint action space, with latent variable approaches inducing consistent temporally extended exploration \citep{mahajan2019maven}. Applications span autonomous driving, unmanned aerial systems, warehousing, and multi-robot teams, illustrating the promise of MARL while underscoring the need for methods that retain coordination efficiency and robustness at real scale \citep{weiss1999multiagent,vlassis2007concise,canese2021multi,bucsoniu2010multi,Zhang2025-ff,Zhang2025-cl,li2025lemix,li2025mace}.

Unlike prior work that treats cooperation failures with fixed reward shaping \citep{ng1999policy}, static communication rules \citep{sukhbaatar2016learning,foerster2016learning}, or single-channel incentives such as LIO~\citep{10.5555/3495724.3496999}, which usually assume honest participants. \Approach{} makes manipulation the central object of study and control. It unifies two levers (policy manipulation over actions and incentive manipulation over rewards) within one learning framework and couples them with adaptive multi-objective optimization that switches in real time between self-centered reward maximization and disruption-oriented strategies based on observed behavior. This yields a single adversarial formulation that covers varying degrees of hostile dynamics rather than separate variants tuned offline to one operating point.

\subsection{Adversarial Attack against Multi-agent Reinforcement Learning}

Adversarial attacks raise greate attentions of safety concerns of widely used deep learning based systems~\citep{li2023white,chen2025your, chen2022nmtsloth, chen2023dark,chen-etal-2023-dynamic,10852400,chen-etal-2024-unveiling,10637681,chen2022nicgslowdown,fu2024safety}. 
A growing body of work shows that multi-agent RL is richly exposed to attack surfaces: adversarial policies can reliably elicit failures from otherwise strong agents \citep{gleave2019adversarial}; observation/actuation perturbations and poisoning degrade coordination by corrupting states, actions, or rewards over time — now including mixed action-plus-reward attacks tailored to online MARL \citep{liu2023efficient_marl_attack}; communication-channel attacks craft deceptive messages that appear benign yet derail team performance \citep{tu2021adversarial_comm}; and backdoors implant triggers that flip behavior in both competitive and cooperative settings.

Despite this progress, almost all demonstrations occur in simulation (e.g., SMAC skirmishes and PettingZoo suites) with desktop-class training/evaluation loops, not in embedded robots with restricted power, latency, and memory \citep{samvelyan2019starcraft,terry2021pettingzoo,li2023rt}. We find little evidence of end-to-end, on-device attack implementations that account for real sensor stacks, real-time scheduling, and closed-loop control on platforms like NVIDIA Jetson, precisely the conditions under which multi-robot systems are deployed \citep{jetson_orin_nano}. This gap motivates manipulation-aware MARL that is benchmarked beyond simulators and stress-tested under embedded constraints (compute budgets, thermal throttling, communication bandwidth), moving the security discussion from ``in-silico” proofs-of-concept to realistic autonomy \cite{Ghinani2025-pf}.

\subsection{Energy harvest and battery longevity for multi-robot systems}

Power and energy consumption are first-order concerns for multi-robot systems, especially at the edge where onboard compute runs under tight power and thermal budgets. From the systems side, datacenter-class advances in warm-water cooling point to practical ways to manage heat with high efficiency: fine-grained warm-water loops improve economy by targeting component-level hotspots~\cite{jiang2019fine}, while hotspot-relievable designs adapt these ideas to energy-efficient edge facilities that often host robotic backends~\cite{pei2022cooledge}. Beyond removing heat, recycling it matters: warm-water setups can harvest and repurpose waste heat to electricity, improving the end-to-end energy picture~\cite{zhu2020heat}; a Dynamic ambient environment may affect harvesting rate, leaving sub-optimal operations on a single agent~\cite{zhang2024mii}. In parallel, power-aware communication stacks reduce the energy cost of coordination and learning—e.g., collective communication that explicitly accounts for power budgets during training and distributed operation~\cite{jia2024pccl}.

At the mission/planning layer, recent top-venue results address energy-aware coordination under charging and endurance constraints. In persistent or long-horizon deployments, robots must interleave task progress with returns-to-charge; approximately optimal assistance scheduling and supervisor routing provide strong baselines for coordinating limited energy across teams~\cite{ji2022traversing}. Fielded multi-modal teams illustrate how exploration policies and duty-cycling are co-designed with endurance, yielding efficient coverage in energy-constrained subterranean settings~\cite{kulkarni2022autonomous,zhang2024mii}. On the hardware/power side, integrated harvesting and hibernation can materially extend endurance, e.g., solar-assisted VTOL platforms that opportunistically accumulate charge between sorties~\cite{carlson2022integrated}.

These mechanisms align with widely adopted efficiency principles and metrics. Energy-proportional computing argues that platforms should draw power roughly in proportion to load, a goal that guides both robot compute selection and scheduling policies~\cite{barroso2007energy}. Facility- and enclosure-level practices are standardized via PUE and thermal guidelines, which provide actionable targets for power distribution and safe operating envelopes when robots rely on edge micro-datacenters or shared lab clusters~\cite{thegreengrid2012pue,ashrae2015thermal}. Taken together, cooling and heat recycling in the backend, power-aware communication, energy-aware planning with charging/harvesting, and energy-proportional operation form a coherent toolkit for sustaining multi-robot systems under real-world power and thermal constraints.

\section{Conclusion}
We introduced \Approach{}, a framework that manipulates inter-robot communication and action selection via \emph{incentive} and \emph{policy} levers and, crucially, \emph{adapts} their balance with a Pareto-aware multi-objective controller. Experiments in Gazebo on ER and IPD show that \Approach{} can exploit non-stationarity to (i) skew rewards toward a chosen adversarial robot, or (ii) entirely depress team success under active disruption. A real Jetson Orin Nano case study will illuminate the systems costs of deploying these mechanisms in the wild.

\section{Ethical Statement}
\label{sec:ethics}
This study examines manipulation mechanisms in multi-robot reinforcement learning with the primary goals of stress-testing robustness and exposing vulnerabilities in cooperative environments. To keep the work firmly within a safe and scientific scope, all adversarial evaluations were conducted either in simulation (Gazebo) or on a fenced, non-safety-critical lab setup with small robots operating under hard emergency stops, speed limits, and continuous supervision. No humans or animals were research subjects, no personally identifiable or sensitive data were involved, and no third-party infrastructure was affected; under institutional guidelines, human-subjects review was not required.

Because manipulation of incentive and policy channels could, if applied irresponsibly, reduce task success or safety margins, bias resource allocation, or conceal harmful behavior by stalling exploration, we designed and recommend layered safeguards. In practice, this means bounding incentives with per-step and per-episode budgets and saturation checks; maintaining an exploration component that is independent of received incentives; authenticating and integrity-protecting inter-robot messages with sequence numbers and replay defenses; recording policies, incentive transfers, and critical actions to a tamper-evident log that supports online and post-hoc audit; constraining the action space near people and obstacles with rule-based safety supervisors and physical e-stops; and tracking health signals such as welfare trends, incentive variance, and gradient-conflict indicators to trigger safe fallbacks to cooperative baselines. These measures are complemented by staged deployment that proceeds from simulation to sandbox to tightly monitored pilots, together with hazard analyses and clear rollback plans, so that insights from controlled testing transition smoothly into practice without compromising safety.

Finally, to reduce misuse risk beyond the lab, we do not release pre-tuned adversarial weights. In sum, by making manipulation dynamics measurable and controllable, this work aims to support safer MARL systems through better testing, detection, and defense design; at the same time, we explicitly discourage deployment of adversarial modes outside controlled research settings and remind practitioners that adoption must comply with applicable laws, standards, and organizational safety policies comapred to exisiting methods~\citep{yang2021adaptive,zhou2024reciprocal,yu2024robust}.

\bibliographystyle{plainnat}
\bibliography{ref}

\end{document}